\documentclass[letterpaper, 10 pt, conference]{ieeeconf}  

\IEEEoverridecommandlockouts                              

\usepackage{graphicx}
\usepackage{subfig}
\usepackage{amsmath,amssymb} 
\usepackage{color}
\usepackage{tabularx}
\usepackage{amsmath}
\usepackage{cite}
\usepackage{lipsum}
\usepackage{mathtools}
\usepackage{hyperref}
\usepackage[percent]{overpic}
\usepackage{mathrsfs}
\usepackage{tikz}
\usepackage{xparse}

\usepackage{mwe}
\usepackage{times}

\tikzset{
  annotatedImage/x/.initial = 0.7,
  annotatedImage/y/.initial = 0.7,
  annotatedImage/width/.initial = 1,
  annotatedImage/.unknown/.code = {
    \edef\tikzappend{\noexpand\tikzset{annotatedImage/.append style =
                {\pgfkeyscurrentname=\pgfkeyscurrentvalue}}}
    \tikzappend
  },
  annotatedImage/.style = {
  }
}

\newsavebox\annotatedImageBox

\newcommand\AnnotatedImageVal[1]{\pgfkeysvalueof{/tikz/annotatedImage/#1}}
\newcommand\SetUpAnnotatedImage[2]{
    \tikzset{annotatedImage/.cd, #1}%
    \sbox\annotatedImageBox{\includegraphics[width=\AnnotatedImageVal{width}\textwidth,
                                          keepaspectratio]{#2}}%
    \pgfmathsetmacro\annotatedHeight{\ht\annotatedImageBox/28.453}
    \pgfmathsetmacro\annotatedWidth{\wd\annotatedImageBox/28.453}%
}

\NewDocumentCommand\annotatedImage{ O{} m m}{%
  \bgroup
    \SetUpAnnotatedImage{#1}{#2}%
    \begin{tikzpicture}[xscale=\annotatedWidth, yscale=\annotatedHeight]%
        \node[inner sep=0, anchor=south west] (image) at (0,0) {\usebox{\annotatedImageBox}};
        \node[annotatedImage] at (\AnnotatedImageVal{x},\AnnotatedImageVal{y}) {#3};
    \end{tikzpicture}
  \egroup%
}

\newcommand\annotate[1][]{\node[annotatedImage,#1]}

\newenvironment{AnnotatedImage}[2][1]{%
  \SetUpAnnotatedImage{#1}{#2}%
  \tikzpicture[xscale=\annotatedWidth, yscale=\annotatedHeight]
    \node[inner sep=0, anchor=south west,inner sep=0] at (0,0) {\usebox{\annotatedImageBox}};
}{\endtikzpicture}

\newcommand{\bx}{\mathbf{x}}

\newcommand{\by}{\mathbf{y}}

\newcommand{\baa}{\mathbf{a}_o}
\newcommand{\bab}{\mathbf{a}_s}

\newcommand{\ioua}{$\mathrm{IoU}_\mathrm{0.25}$}
\newcommand{\ioub}{$\mathrm{IoU}_\mathrm{0.5}$}
\newcommand{\iouc}{$\mathrm{IoU}_\mathrm{0.75}$}
\DeclareMathOperator*{\argmin}{argmin} 

\definecolor{dgreen}{rgb}{0,0,0}
\definecolor{dyellow}{rgb}{.7,.7,0}
\definecolor{dred}{rgb}{1,0,0}
\definecolor{dblue}{rgb}{0,0,0.7}
\definecolor{dorange}{rgb}{0.9,0.5,0.1}

\title{\LARGE \bf
Learning Object Placements For Relational Instructions \\by Hallucinating Scene Representations 
}

\author{Oier Mees*, Alp Emek*, Johan Vertens, Wolfram Burgard
\thanks{$^\ast$These authors contributed equally. All authors are with the University of Freiburg, Germany. Wolfram Burgard is also with the Toyota Research Institute, Los Altos, USA. This work has  been supported partly by the Freiburg Graduate School of Robotics and the German Federal Ministry of Education and Research under contract number 01IS18040B-OML.}
}

\begin{document}

\maketitle
\thispagestyle{empty}
\pagestyle{empty}

\begin{abstract}
Robots coexisting with humans in their environment and performing services for them need the ability to interact with them. One particular requirement for such robots is that they are able to understand spatial relations and can place objects in accordance with the spatial relations expressed by their user.
In this work, we present a convolutional neural network for estimating pixelwise object placement probabilities for a set of spatial relations from a single input image. During training, our network receives the learning signal by classifying hallucinated high-level scene representations as an auxiliary task. Unlike previous approaches, our method does not require ground truth data for the pixelwise relational probabilities or 3D models of the objects, which significantly expands the applicability in practical applications.  
Our results obtained using real-world data and human-robot experiments demonstrate the effectiveness of our method in reasoning about the best way to place objects to reproduce a spatial relation. Videos of our experiments can be found at \url{https://youtu.be/zaZkHTWFMKM}

\end{abstract}

\section{INTRODUCTION}

\begin{figure}[t]
\centering
\includegraphics[width=0.9\linewidth]{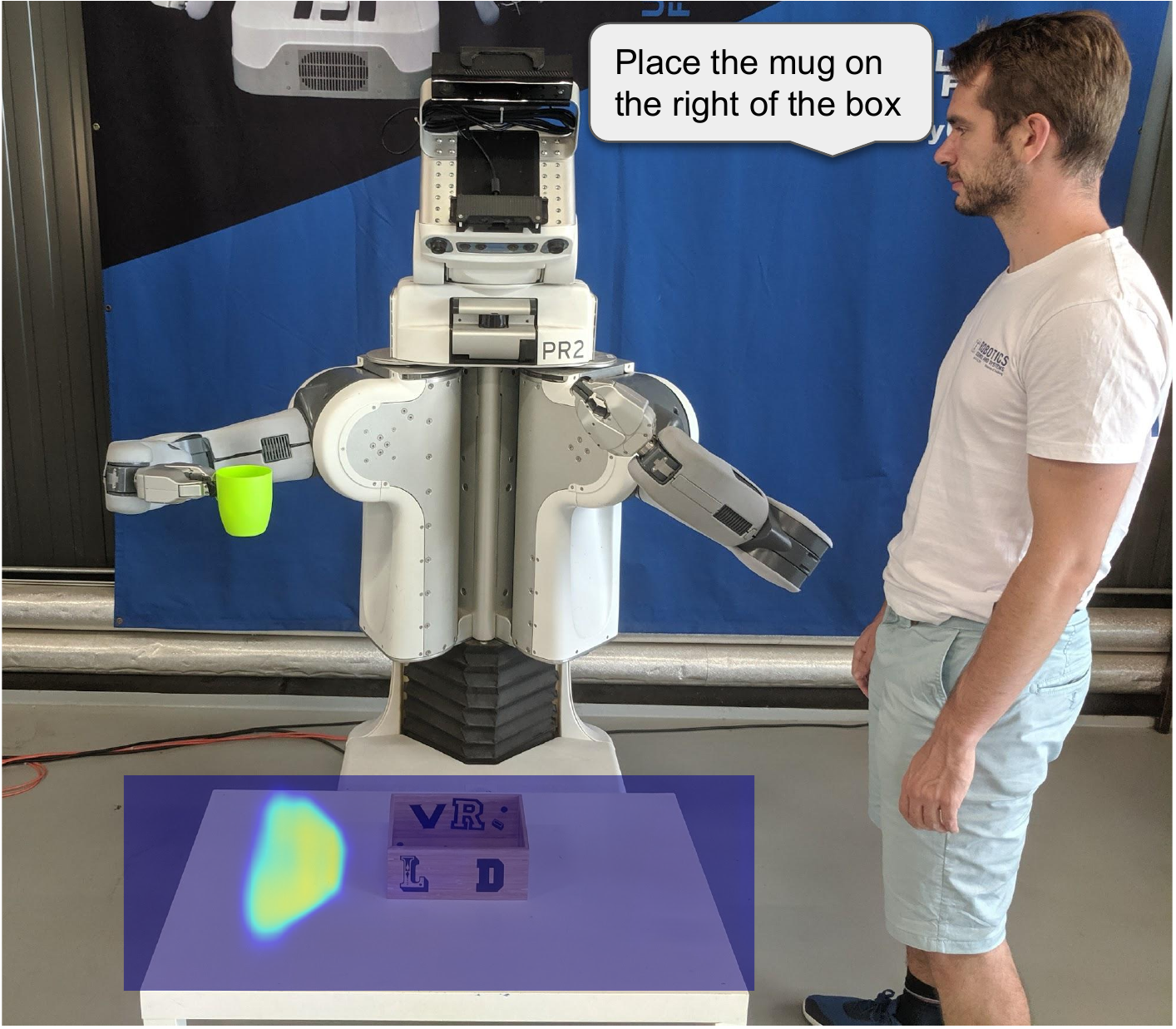}
   \caption{The goal of our work is to follow natural language instructions based on spatial relations to place arbitrary objects. Our network learns to predict pixelwise placing probability distributions (heatmap on the table) solely from classifying hallucinated high-level scene representations into a set of spatial relations. }
\label{fig:teaser}
\end{figure}

Understanding and leveraging spatial relations is a key capability of autonomous service robots operating in human-centered environments. In this work, we aim to develop an approach that enables a robot to understand spatial relations in natural language instructions to place arbitrary objects.  The spatial relations include common ones such as ``left'', ``right'' or ``inside''. In Figure~\ref{fig:teaser},  the robot is asked to ``place the mug on the right of the box''. To do so, the robot needs to reason about where to place the mug relative to the box in order to reproduce said spatial relation. Moreover, as natural language placement instructions do not uniquely identify a location in a scene, it is desirable to model this using distributions to capture the inherent ambiguity.


Object-object spatial relations can be learned  in a fully-supervised manner~\cite{jiang2012learning, jiang2012learning2,mees17iros, jund2018optimization, zampogiannis2015learning, fichtl2014learning, rosman2011learning} from 3D vision. The main limiting factor for exploiting this setup in practical robotics applications is the need for collections of corresponding 3D object shapes and relational data, which are difficult to obtain and require additional instrumentation for object tracking. This limits prior methods to training on synthetic datasets or simulators,  leading to  difficulties in their application to real-world scenarios. A possible solution to this problem is to model relations directly from RGB images~\cite{dai2017detecting}, which  allows direct training on real image data without the need of modeling the scene in 3D. Reasoning about object placements for relational instructions in this context requires estimating  pixelwise spatial distributions of placement locations, as shown in Figure~\ref{fig:teaser}. One of the key challenges to estimate such pixelwise spatial distributions is the lack of ground-truth data.  This originates from the inherent ambiguity on modeling such ground-truth distributions without using heuristics. If one wants to model the relation ``left'', how far left of the reference object would form a valid relation? Should the distribution have a single or multiple modes? And where is the boundary between ``left'' and ``in front of'' for instance? 


In this paper, we push the limits of relational learning further and present a method which leverages a weaker form of supervision to model object placement locations conditioned on a set of spatial relations.  
We address the problem of the unavailability of ground-truth pixelwise annotations of spatial relations from the perspective of auxiliary learning.
Our approach relies solely on relational bounding box annotations and the image context to learn pixelwise  distributions of object placement locations over spatial relations, without any additional form of supervision or instrumentation. Though classifying two objects into a spatial relation does not carry any information on the best placement location to reproduce a relation, inserting objects at different locations in the image would allow to infer a distribution over relations. Most commonly, ``pasting'' objects realistically in an image requires either access to 3D models and silhouettes~\cite{dwibedi2017cut,su2015render,georgakis2017synthesizing,tremblay2018training} or carefully designing the optimization procedure of generative adversarial networks~\cite{lee2018context, lin2018st}. Moreover, naively ``pasting'' object masks in images creates subtle pixel artifacts that lead to noticeably different features and to the training erroneously focusing on these discrepancies. Our results show that such models lead to reduced performance. Instead, we take a different approach and implant high-level features of objects into feature maps of the scene generated by a network to hallucinate scene representations, which are then classified as an auxiliary task to get the learning signal. Training a network in this setup solely  requires being able to classify relations between pairs of objects from an image.

We demonstrate both qualitatively and quantitatively that
our network trained on real-world images successfully predicts pixelwise placement probability distributions for each spatial relation.  Our approach can be trained on images with relational bounding box annotations and does not require 3D information or any additional instrumentation to predict the spatial distribution of arbitrary objects, thus making it readily applicable in a variety of practical robotics
scenarios. We exemplify this by using the probability distributions produced by our method in a robot experiment to place objects on a tabletop scene by following natural language instructions from humans.


\section{Related work}
Learning spatial relations by relying on the geometries of objects provides a robot with the necessary capability to carry out tasks that require understanding object interactions, such as object placing~\cite{jiang2012learning, jiang2012learning2}, human robot interaction~\cite{schulz2017collaborative, guadarrama2013grounding, Shridhar-RSS-18, aly2018towards}, object manipulation~\cite{zampogiannis2015learning} or generalizing spatial  relations to new objects~\cite{mees17iros, jund2018optimization, li2016learning}. Commonly, spatial relations are modeled based on the geometries of objects given their point cloud models~\cite{zampogiannis2015learning,mees17iros,jund2018optimization}.
However, learning object relations from 3D data~\cite{fichtl2014learning, zampogiannis2015learning, mees17iros, rosman2011learning, jund2018optimization} typically requires additional instrumentation to track objects, with the consequent difficulties due to occlusions for example. One way to overcome this limitation could be learning to predict 3D shapes from single images in a self-supervised manner~\cite{mees19iros}.
In contrast to these works, we  learn spatial distributions directly from real-world  images.

Spatial relations also play a crucial role in understanding natural language instructions~\cite{paul2016efficient,hatori2018interactively,magassouba2018multimodal}, as objects are often described in relation to others. Several studies on human–robot interactions have been conducted, mainly for picking objects. These works focus on analyzing the expressive space of abstract spatial concepts as well as notions of ordinality and cardinality ~\cite{Shridhar-RSS-18, paul2016efficient}. Complementary to these works, we propose to learn distributions of object  relations,  for commonly used prepositions in natural language, to enable a service robot to place arbitrary objects given natural language instructions.

There has been a large body of research targeting relations in the vision community.
Multiple works attempt to ground object relationships from images for classification~\cite{dai2017detecting}, referring expression comprehension~\cite{nagaraja2016modeling,yu2018mattnet,krishna2018referring}, human-object interactions~\cite{gkioxari2018detecting} or relational learning in visual question answering~\cite{battaglia2016interaction, santoro2017simple}. While these approaches reason about an existing scene, our method learns which future state might follow best a spatial relation grounded in natural language instructions. Generating a future state in a object placement scenario would mean to insert the object to be placed into different locations in the robot's view image. There exists a plethora of work for learning how to synthetize objects realistically into images~\cite{dwibedi2017cut,su2015render,georgakis2017synthesizing,tremblay2018training}. Most commonly, such methods  requires either access to 3D models and silhouettes  or carefully designing the optimization procedure of generative adversarial networks~\cite{lee2018context, lin2018st}. Instead, our approach implants high-level features of objects to hallucinate scene representations, which are then classified by an auxiliary network to learn spatial distributions. 

\begin{figure}[t]
\vspace*{-2mm}
\subfloat[][]{\includegraphics[width=0.5\linewidth]{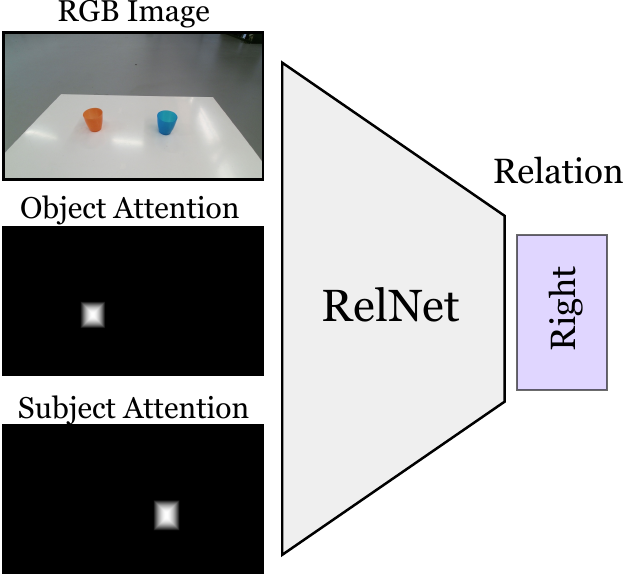}\label{fig:standard_relnet}
}
\subfloat[][]{\includegraphics[width=0.5\linewidth]{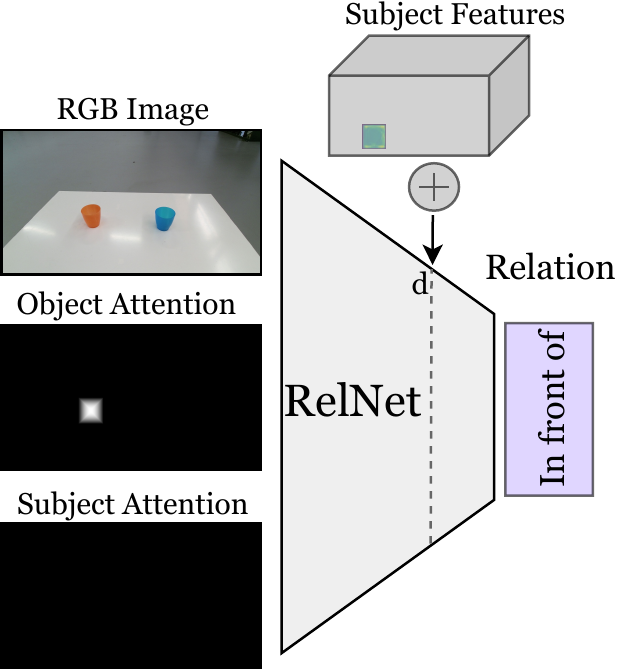}\label{fig:implanting_motivation}
}    
\caption{In the first stage of our approach, we train an auxiliary convolutional neural network, called RelNet, to predict spatial relations given the input image and the two attention masks referring to the two objects forming a relation (a). After training, we can ``trick'' the network to classify hallucinated scenes by implanting high-level features of items at different spatial locations (b).} \label{fig:all_relnets}
\end{figure}

\section{Method description}
In this section we describe the technical details of our method for estimating pixelwise object placement probabilities for a set of spatial relations from a single input image. We consider pairwise  relations and express the subject item as being \emph{in relation to} the reference item. We extract subject, object and relation from natural language instructions.

\begin{figure*}[t]
\centering
\begin{AnnotatedImage}[width=0.93]{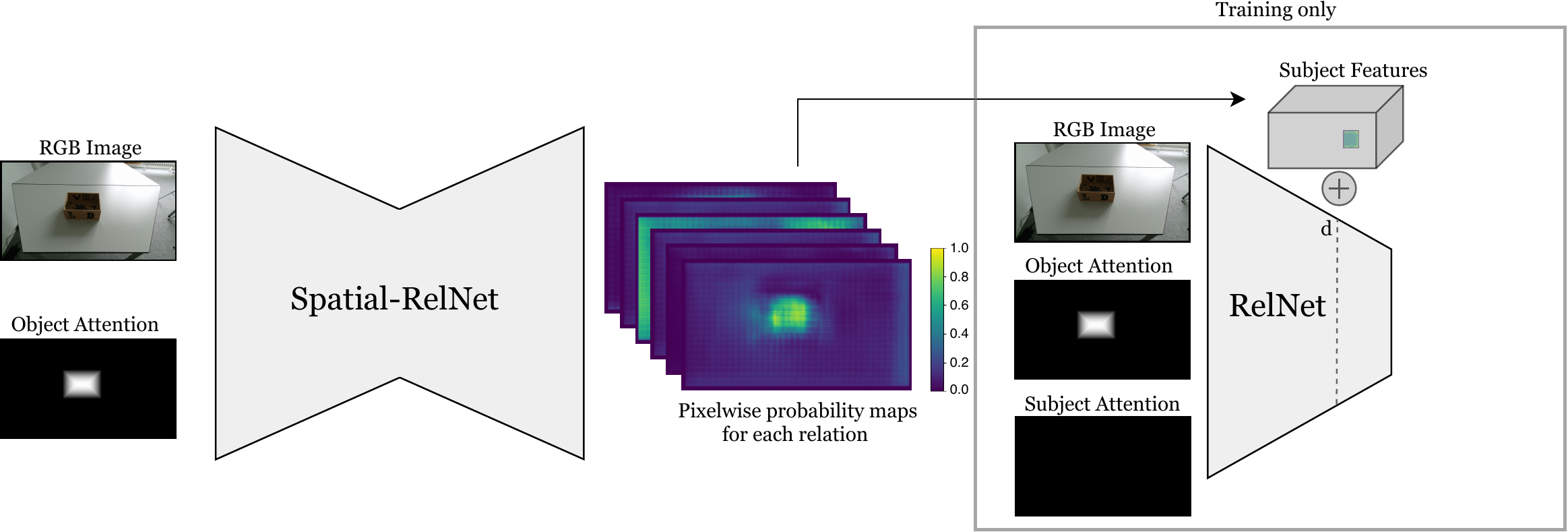}
\annotate (G) at (0.943,0.42){$ \scriptstyle \left\lVert \Gamma(u,v) -  f_{\varphi} \right\rVert_2^2$};
\annotate (G) at (0.57,0.18){$\scriptstyle(\Gamma)$};
\annotate (G) at (0.72,0.85){\small Sample $(u, v)$};
\end{AnnotatedImage}
  \caption{Our encoding-decoding Spatial-RelNet network processes the input RGB image and the object attention mask to produce pixelwise probability maps $\Gamma$ over a set of spatial relations. During training, we sample locations $(u, v)$ according to $\Gamma$, implant inside the auxiliary network RelNet at the sampled locations high level features of objects and classify the hallucinated scene representation to get a learning signal for Spatial-RelNet. At test time the auxiliary network is not used.}
\label{fig:architecture}
\end{figure*}

\subsection{Auxiliary Network}
In the first stage of our approach, we encode the input RGB image  together with an object and a subject attention mask to classify them into a set of spatial relations with an auxiliary convolutional neural network (CNN). We denote the RGB image, the object and subject attention masks as  $\bx^{i},\baa^{i},\bab^{i}$ respectively and $\by^{i}$ corresponds
to the relation label in one-hot encoding -- i.e., $\by^{i}\in\{0, 1\}^{|C|}$ is a vector of dimensionality C (the number of relations). We model relations for a set of commonly used natural language spatial prepositions 
$C= \{\texttt{inside}, \texttt{left}, \texttt{right}, \texttt{in front}, \texttt{behind},  \texttt{on top} \} $.
Let $\mathcal{D} = \lbrace (\bx^{1}, \baa^{1}, \bab^{1}, \by^{1}), \dots,
(\bx^{N}, \baa^{N}, \bab^{N}, \by^{N}) \rbrace$ be the labeled data available for training our auxiliary classification network, which we name RelNet, see Figure~\ref{fig:standard_relnet}. Let $\theta_{RelNet}$ be the parameters of the network.
 We denote the mapping of RelNet as $f(\bx^{i},\baa^{i},\bab^{i};\theta_{RelNet})$ $\in\mathbb{R}^{|C|}$. The attention masks are calculated as a Gaussian distance transform 
$a(u,v) = \frac{1}{\sigma\sqrt{2\pi}} e^{ -\frac{1}{2}\left((1-d_{uv})/\sigma\right)^2 }$ with $d_{uv}$ being the distance transform between $(u, v)$ and the bounding box center, based on the L2 norm and with  $\sigma=2$. 
 
The goal of RelNet is to learn classifying scenes of pairwise object relations  by minimizing the cross-entropy ($\mathit{softmax}$) loss. The  $\mathit{softmax}$ function converts a score $z_c$ for class $C$ into a posterior class probability that can be computed as $\mathscr{L}(z_c) = \exp(z_c)/ \sum_{j=1}^{|C|} \exp(z_j)$. Using stochastic gradient descent (SGD) we then optimize:
\begin{equation}
\theta^{*}_{RelNet} \in \argmin_{\theta_{RelNet}}  \sum_{i=1}^{N} {\mathscr{L}(f(\bx^{i},\baa^{i},\bab^{i};\theta_{RelNet}), \by^{i})}. 
\end{equation}
RelNet is only utilized during training time and discarded at inference time.

\subsection{Hallucinating Scene Representations}

Clearly, classifying the spatial relation formed by two items is not suitable to identify the best placing location to reproduce a relation. However, inserting objects at different locations in the image would allow to infer a distribution over relations. As mentioned before, ``pasting'' objects realistically in an image requires commonly either access to 3D models and silhouettes~\cite{dwibedi2017cut, su2015render, georgakis2017synthesizing, tremblay2018training} or carefully designing the optimization procedure of generative adversarial networks~\cite{lee2018context, lin2018st}. To tackle this challenge, we take a different approach and implant high-level features of objects into a high-level feature representation of RelNet to hallucinate scene representations, which are then classified to get the learning signal, as shown in Figure~\ref{fig:implanting_motivation}. 
Given an input image $\bx^{i}$ of size $W \times H$, we use the RelNet network on the image to obtain a spatial feature map  $M_o$ of size $W_f \times H_f \times N_f$  (width, height, number of filters) at  depth $d$. Given an input image, we extract a slice of the feature map $s \in \mathcal{R}^{W_s \times H_s \times N_f}$ corresponding to a bounding box containing an item in the image. Thus, hallucinating a scene representation requires no more than making a forward pass with RelNet  and implanting the high level features of a subject object $s$ into the feature map $M_o$ at a sampled location $(u,v)$ by summation and continuing the forward pass with the modified feature map. We define the implanting operation as:
\begin{equation}
\varphi(M_o, s, u, v) =  M_o + M_s(u,v), 
\end{equation}
where 
\begin{equation}
(M_s(u, v))_{jk} = \begin{cases}
    s(j-u, k-v), & \parbox{3cm}{if $u\leq j \leq u + W_s$ and $v\leq k \leq v + H_s$}.\\
    0, & \text{otherwise}.
  \end{cases}
\end{equation}
This way we can reason over what pairwise spatial relations are most likely to be formed given an existing item in the image and a subject item which can be hallucinated at different locations. In other words, what relation would the two items form, if the subject item was placed at the specified location. 
Formally, we define the mapping of a RelNet with implanted features $s$ at location $u, v$ as $f_{\varphi}(\bx^{i},\baa^{i},\bab^{i}, s, u, v)\in\mathbb{R}^{|C|}$.

\subsection{Learning pixelwise item placement distributions}
In the final stage of our approach, we model the primary task of inferring pixelwise spatial distributions to find the best placing locations by following a natural language instruction containing a spatial relation. We define a second network, named Spatial-RelNet, with an encoding-decoding architecture. Given an  image $\bx^{i}$ of size $W \times H$ and the object attention mask $\baa^{i}$, the network predicts for each pixel in the input image the probabilities of belonging to one of the $C$ classes with respect to the reference object attention, see Figure~\ref{fig:architecture}. Thus, we denote the mapping of Spatial-RelNet as $g(\bx^{i},\baa^{i}) =  \Gamma \in\mathbb{R}^{W \times H \times |C|}$ and $\sum_{j=1}^{|C|} \Gamma_j(u,v)=1$ for all $u = 1...W, v = 1...H$. Due to the unavailability of ground-truth pixelwise annotation of spatial relations we propose a novel formulation which leverages auxiliary learning.
During training, we sample pixel locations $(u,v) \in \xi \subseteq \{0,...,W\} \times \{0,...,H\}$ according to the probability maps $\Gamma$ produced by Spatial-RelNet and then implant at the specified locations the subject object features to compute with RelNet a posterior class probability over relations. This way, we can reason over what relation would most likely be formed if we placed an object at the given location. Our formulation allows predicting non-parametric probability distributions.
Thus, by sampling multiple locations in the scene from $\Gamma$ and leveraging the auxiliary task of classifying the spatial relation formed by two objects in an image we can train our primary network Spatial-RelNet with the following mean squared error loss:
\begin{equation}
\sum_{u,v \in \xi} ||{g(\bx^{i},\baa^{i})_{uv}  - f_{\varphi}(\bx^{i},\baa^{i},\bab^{i}, s, u, v)}||_2^2. 
\end{equation}
We note that at inference time the auxiliary RelNet network is discarded.

\subsection{Implementation Details}
In the first stage of our approach, we train the auxiliary RelNet network on the task of classifying spatial relations between two objects in images. The auxiliary RelNet network is based on a ResNet-18~\cite{he2016deep} architecture and is initialized with ImageNet pre-trained weights. We use the SGD optimizer with a learning rate of $10^{-3}$.


In the final stage of our approach, we train Spatial-RelNet to predict pixelwise  spatial distributions by using the auxiliary RelNet network for supervision. The Spatial-RelNet is inspired by the FastSCN~\cite{poudel2019fast} semantic segmentation architecture and initialized randomly. We apply a per-pixel sigmoid activation function for the last layer instead of $\mathit{softmax}$.   We  use the ADAM optimizer with  a learning rate of $10^{-3}$.
 We sample 20 locations per distributions and use a feature map of the size $128, 10, 10$ pertaining to an object to hallucinate the scene representations. For all experiments, we implant the subject features after the third convolutional block ``conv3\_x'' ($d=3$) of the ResNet-18 architecture that characterizes RelNet. In order to speed up the  training we apply a Sobel filter on the output probability maps $\mathbf{H}$ to propagate the gradient to local neighborhoods. We define the Sobel kernels as $ \left[\begin{smallmatrix} 
-1 & 0 & 1 \\
-2 & 0 & 2 \\
-1 & 0 & 1
\end{smallmatrix}\right]$ for the $x$ direction and $ \left[\begin{smallmatrix} 
1 & 2 & 1 \\
0 & 0 & 0 \\
-1 & -2 & -1
\end{smallmatrix}\right]$ for the $y$ direction.

\section{Experiments}
In this section we showcase our approach both qualitatively and quantitatively, and demonstrate its applicability in a human-robot experiment, where participants ask a PR2 robot to place objects with natural language instructions based on spatial relations.

\subsection{Dataset}
\begin{figure}[b]
\centering
\setlength{\tabcolsep}{0.5pt}
\subfloat[][]{\includegraphics[width=0.32\linewidth]{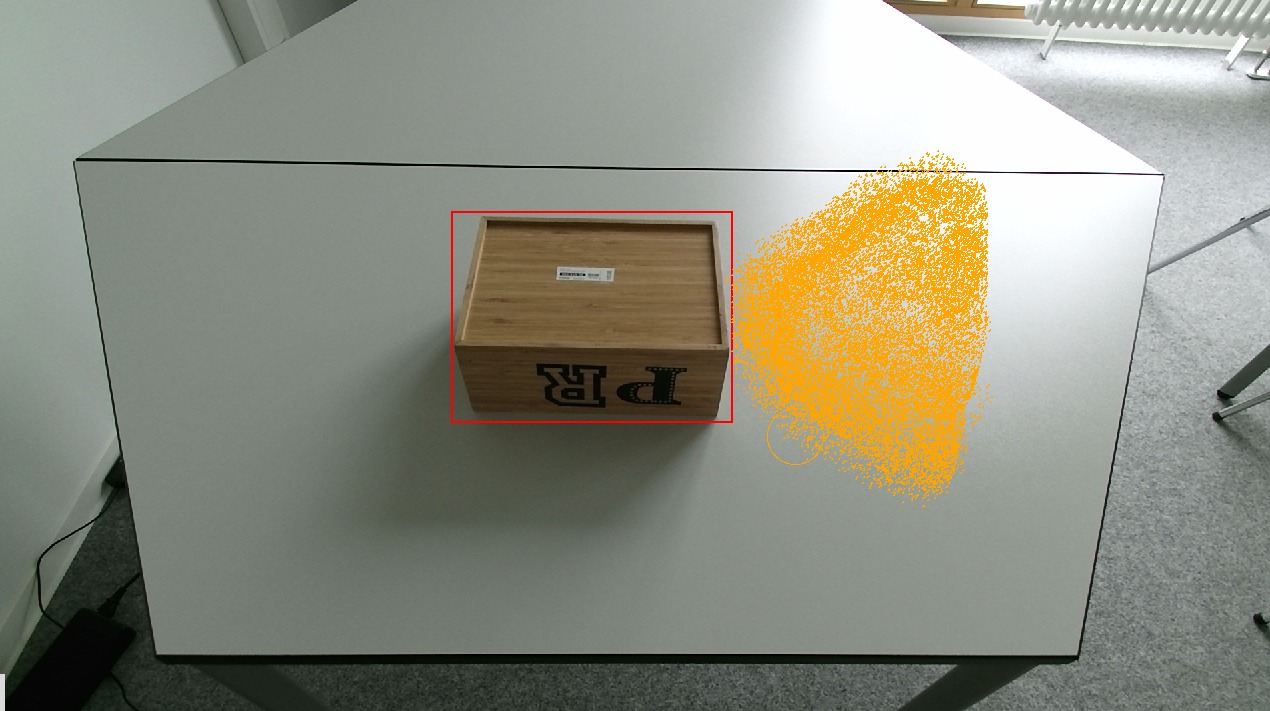}\label{fig:spraya}
}
\subfloat[][]{\includegraphics[width=0.32\linewidth]{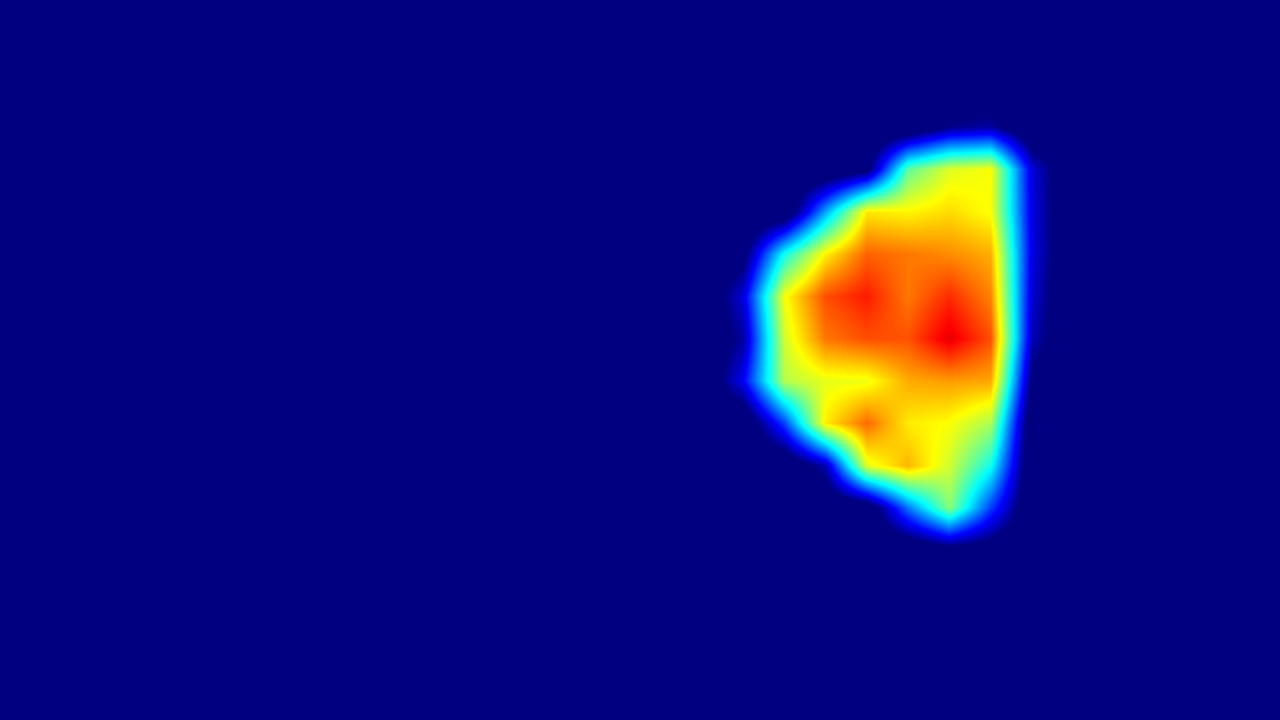}\label{fig:sprayb}
}
\subfloat[][]{\includegraphics[width=0.32\linewidth]{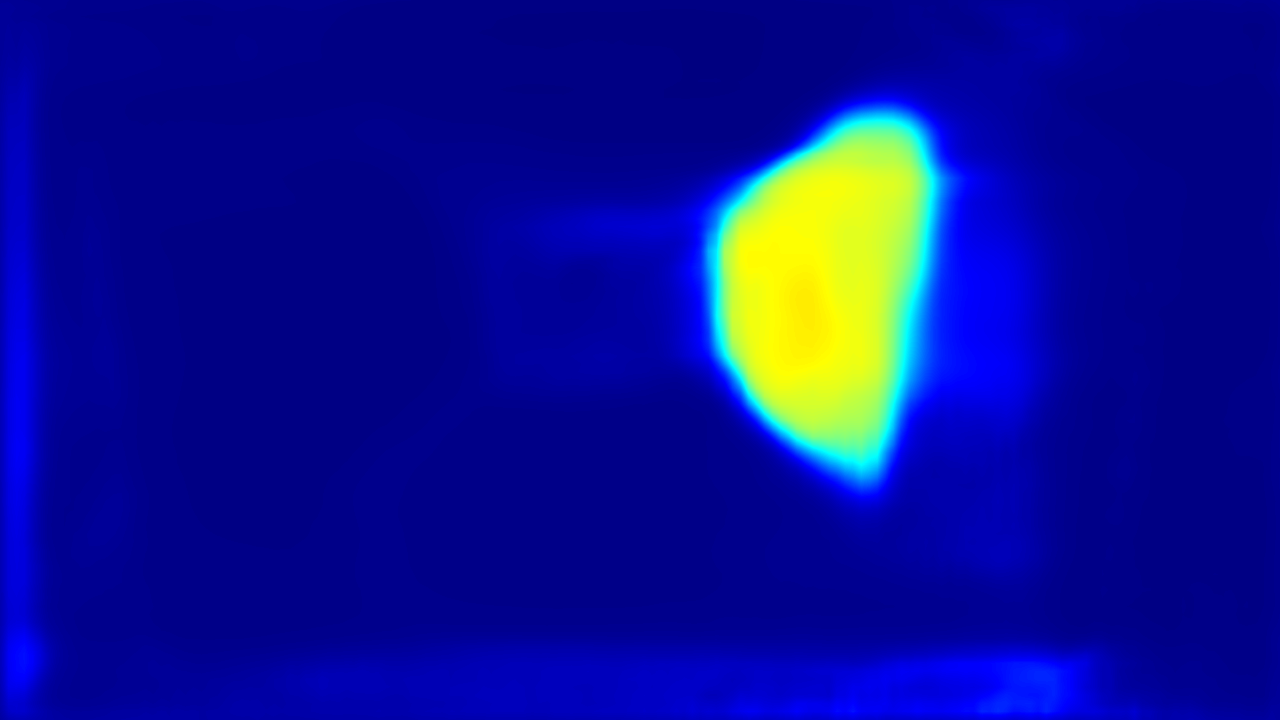}
}    
\caption{Example user annotation of ground-truth points for the relation ``right'' with a ``spray'' paint tool (a). We convolve the user annotated points to generate a dense distribution (b). The shown  network output distribution and the  ground-truth distribution have a \ioub ~of 0.39 (c).} \label{fig:annottaion}
\end{figure}

We record and annotate a total of 1237 images of 165 tabletop scenes. The images depict tabletop scenes from three different viewpoints (object-centric to top-down) containing spatial relations formed by using combinations of 40 different household objects. Learning from multiple camera viewpoints helps the approach become less sensitive to viewpoint changes and generalize better. We annotate 5304 pairwise bounding boxes with the  commonly used natural language spatial prepositions 
$C= \{\texttt{inside}, \texttt{left}, \texttt{right}, \texttt{in front}, \texttt{behind},  \texttt{on top} \} $.  For all recorded scenes we use different tablecloths and tables in different rooms to avoid overfitting. To evaluate the pixelwise probability distributions predicted by Spatial-RelNet, we record  105 scenes containing unseen objects and tables. Due to the inherent ambiguity of defining pixelwise spatial distributions, we ask 3 participants to annotate them. To do so, the user use a ``spray'' tool to draw points in which placing an item would reproduce a given spatial relation, as shown in Figure~\ref{fig:spraya}. The points are then convolved with a fixed kernel to generate a dense pixelwise ground-truth distribution, as seen in Figure~\ref{fig:sprayb}.
\begin{figure}[t!]
\centering
\setlength{\tabcolsep}{2pt}
\begin{tabular}{ccc}
\vspace*{0.35mm}
\includegraphics[width=0.3\linewidth, trim=100 20 500 20, clip]{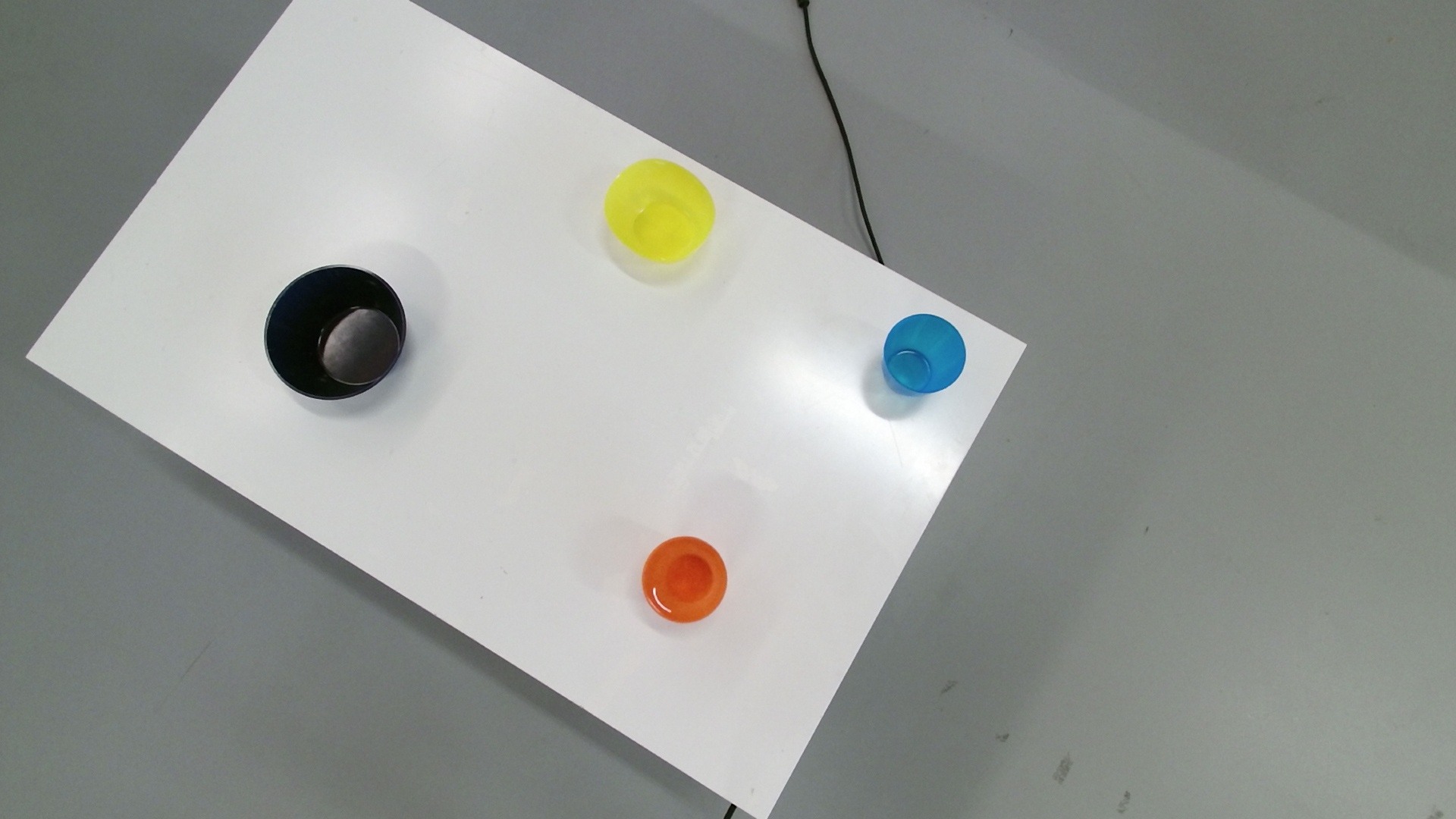}&
\includegraphics[width=0.3\linewidth,  trim=100 20 500 20, clip]{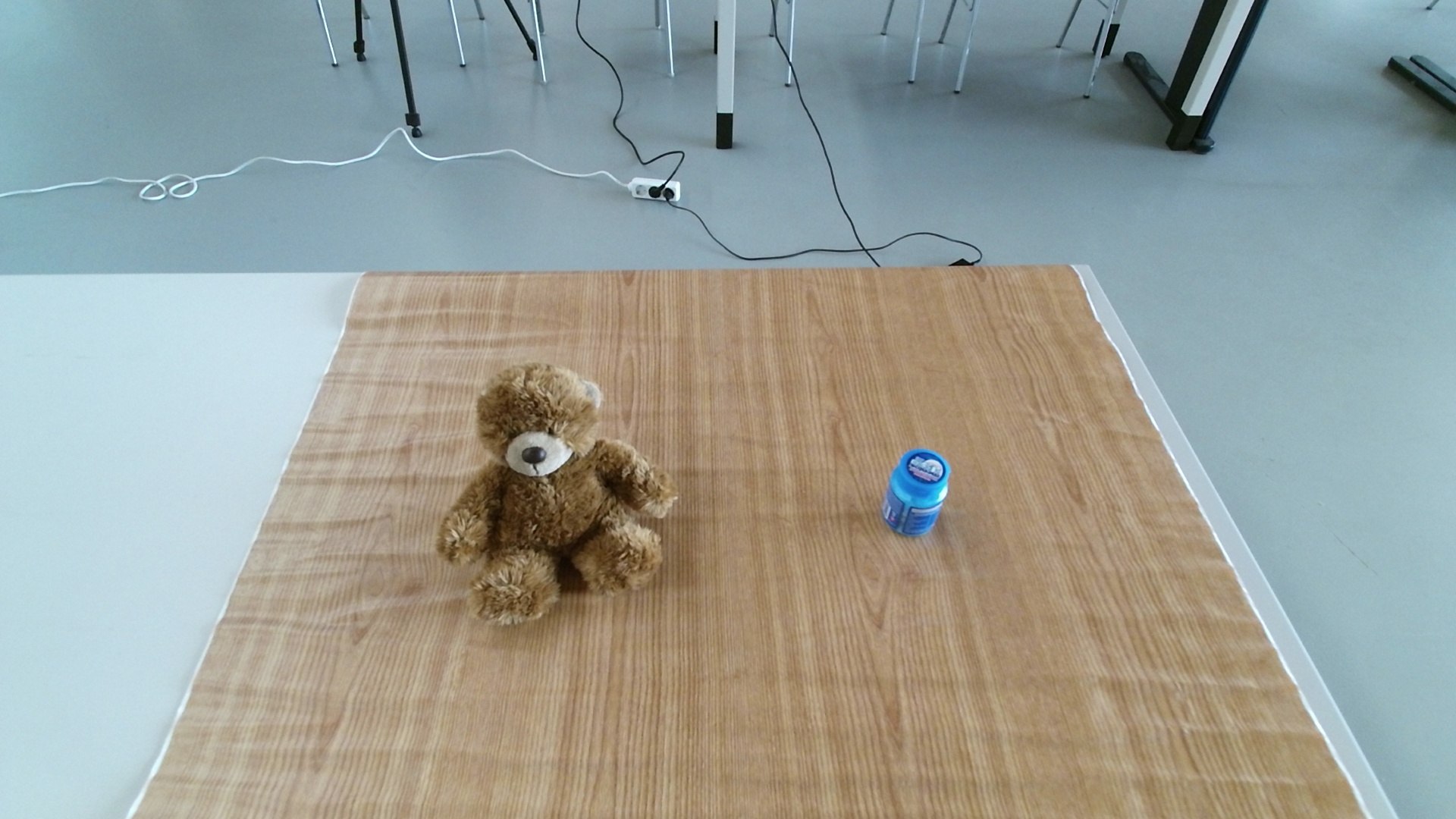}&
\includegraphics[width=0.3\linewidth,  trim=300 20 300 20, clip]{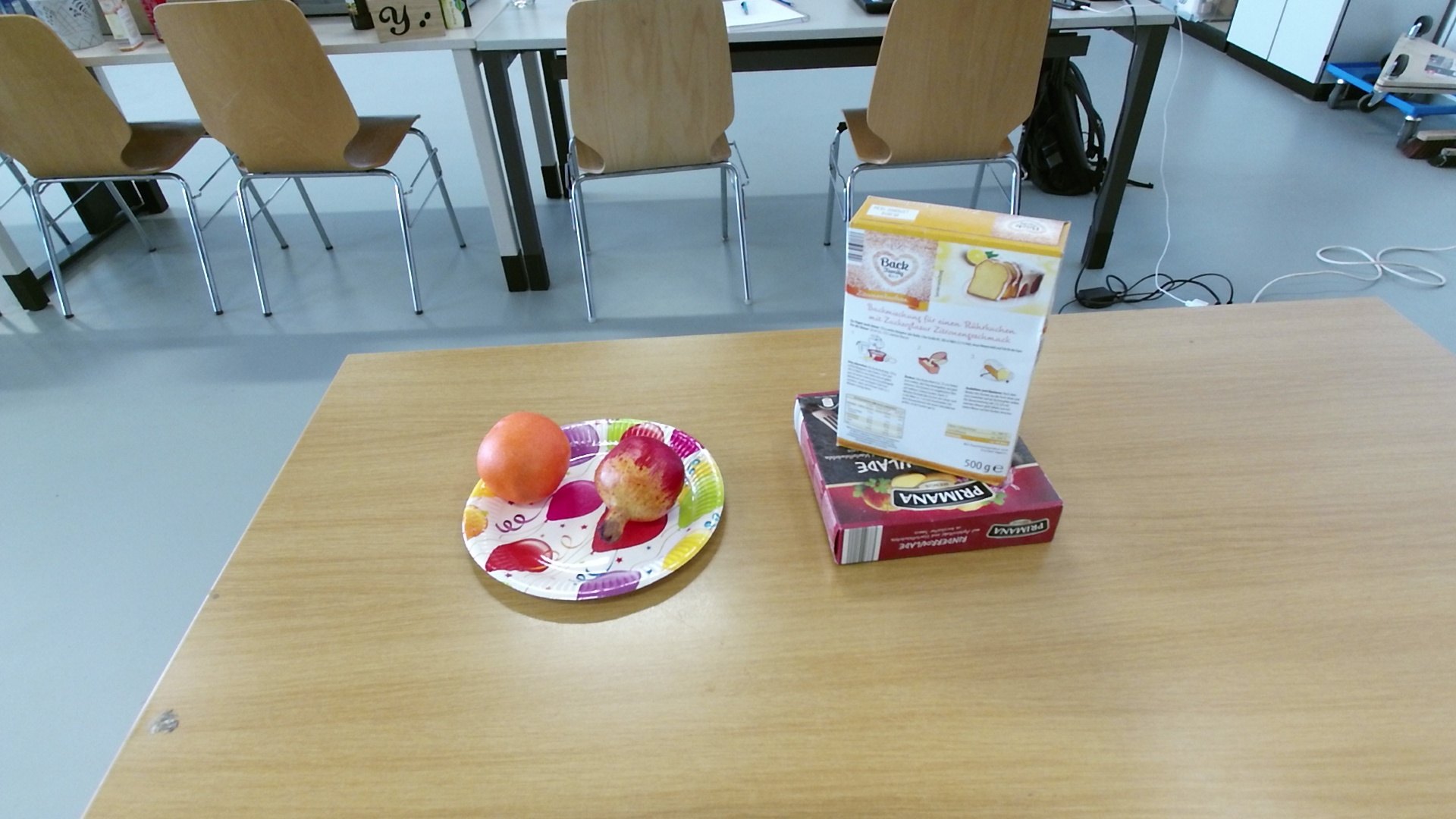}\\
\includegraphics[width=0.3\linewidth,  trim=300 20 300 20, clip]{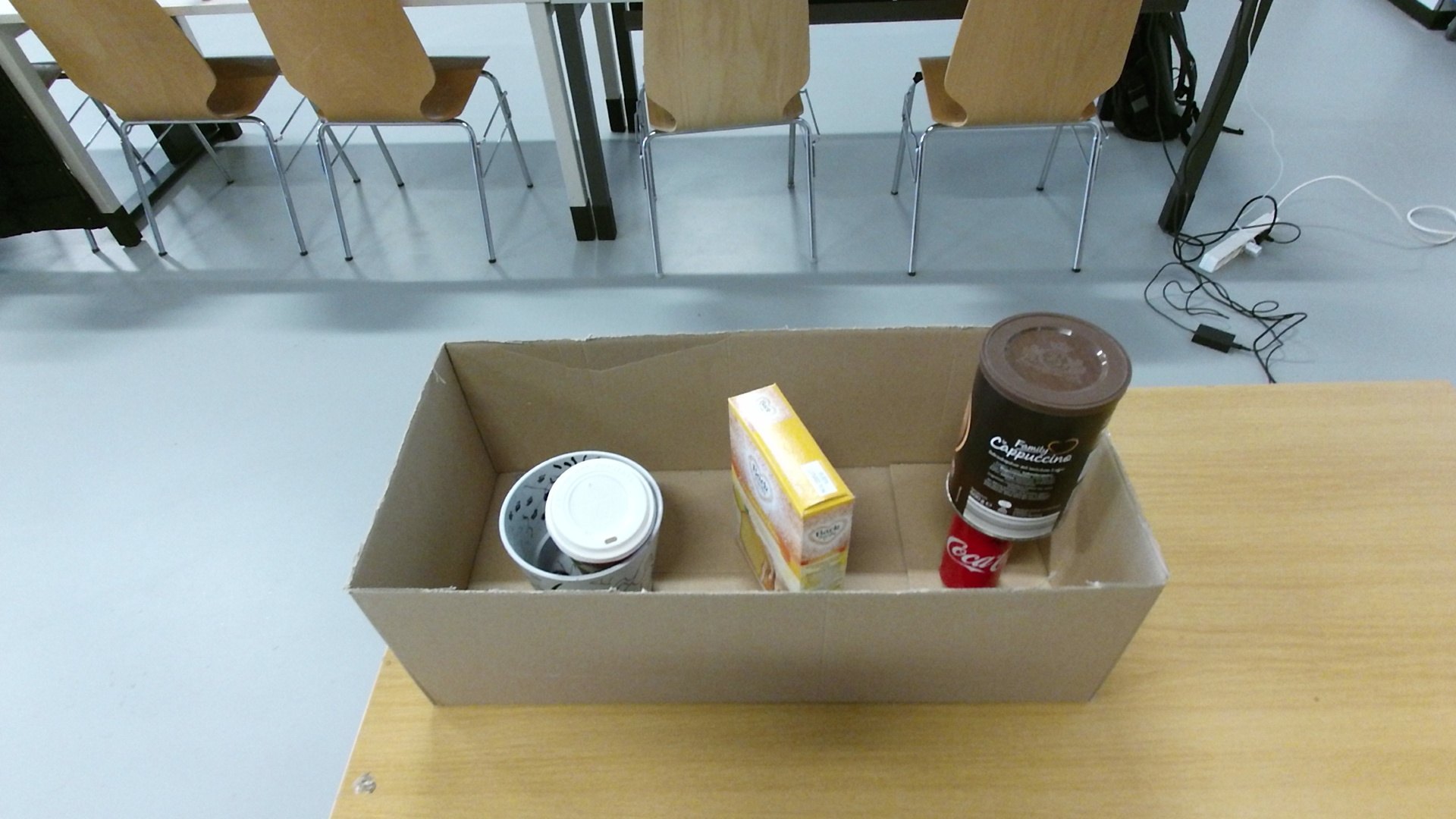}&
\includegraphics[width=0.3\linewidth,  trim=100 20 500 20, clip]{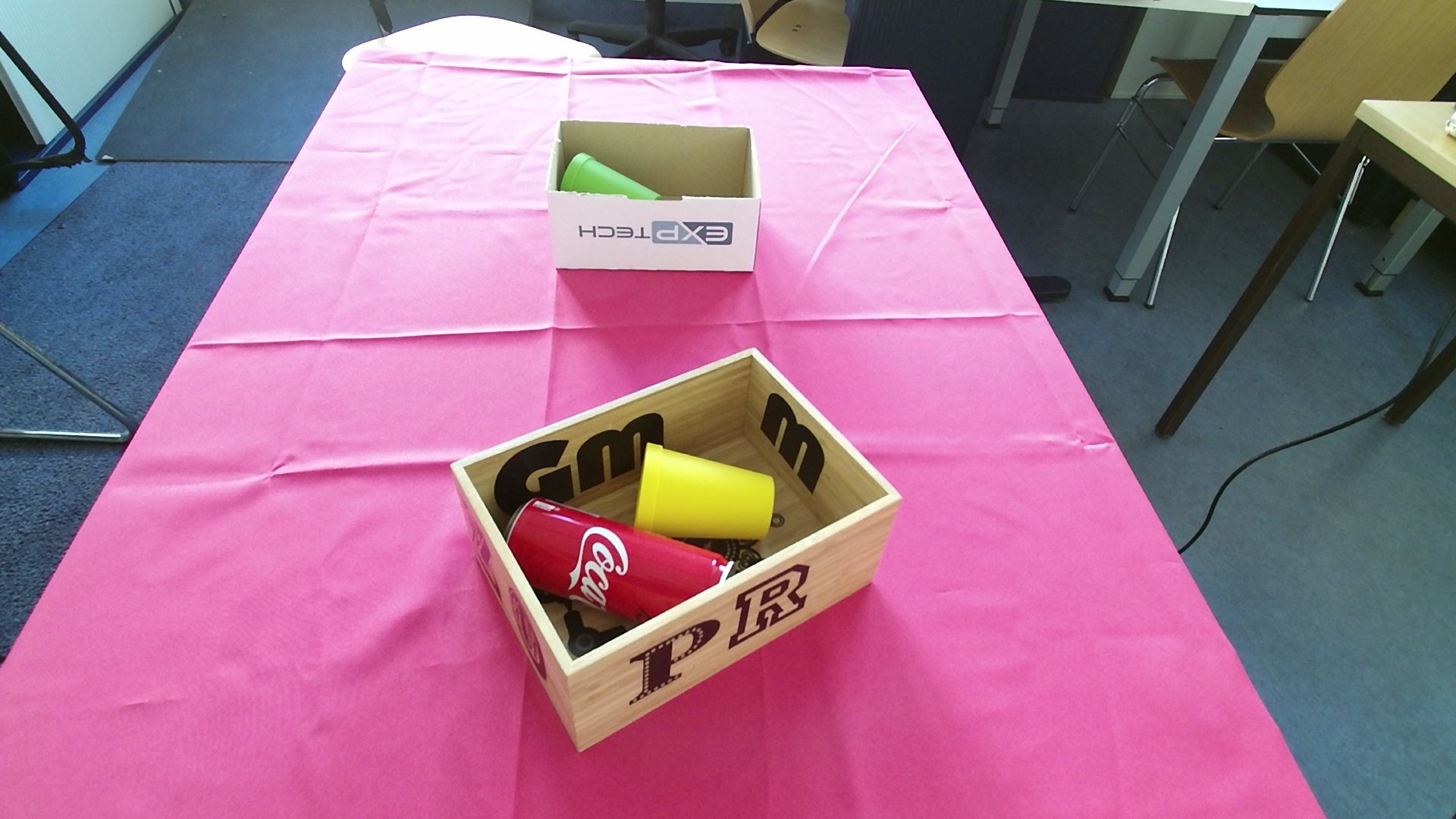}&
\includegraphics[width=0.3\linewidth,  trim=100 20 500 20, clip]{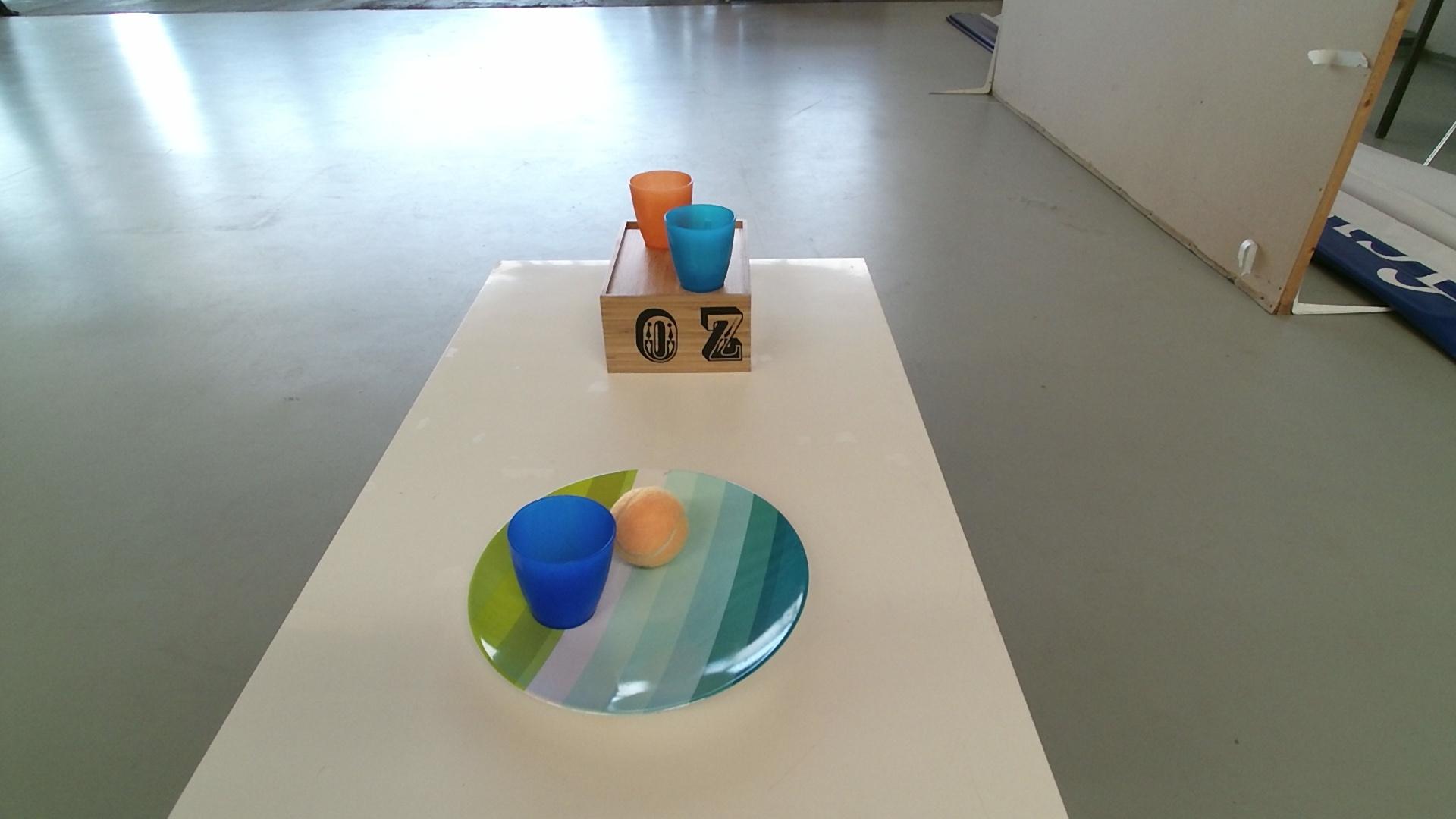}
\end{tabular}

\caption{Examples of the scenes  recorded for training the auxiliary RelNet classifier. We recorded a total of 1237 images of 165 tabletop scenes and manually annotated the bounding boxes of the objects and their  spatial relations.}
\label{fig:dataset}
\end{figure}
\begin{figure*}[t]
\centering
\vspace*{4mm}
\setlength{\tabcolsep}{1.35pt}
\begin{tabular}{c c c c c c c}
\begin{overpic}[width=0.135\linewidth]{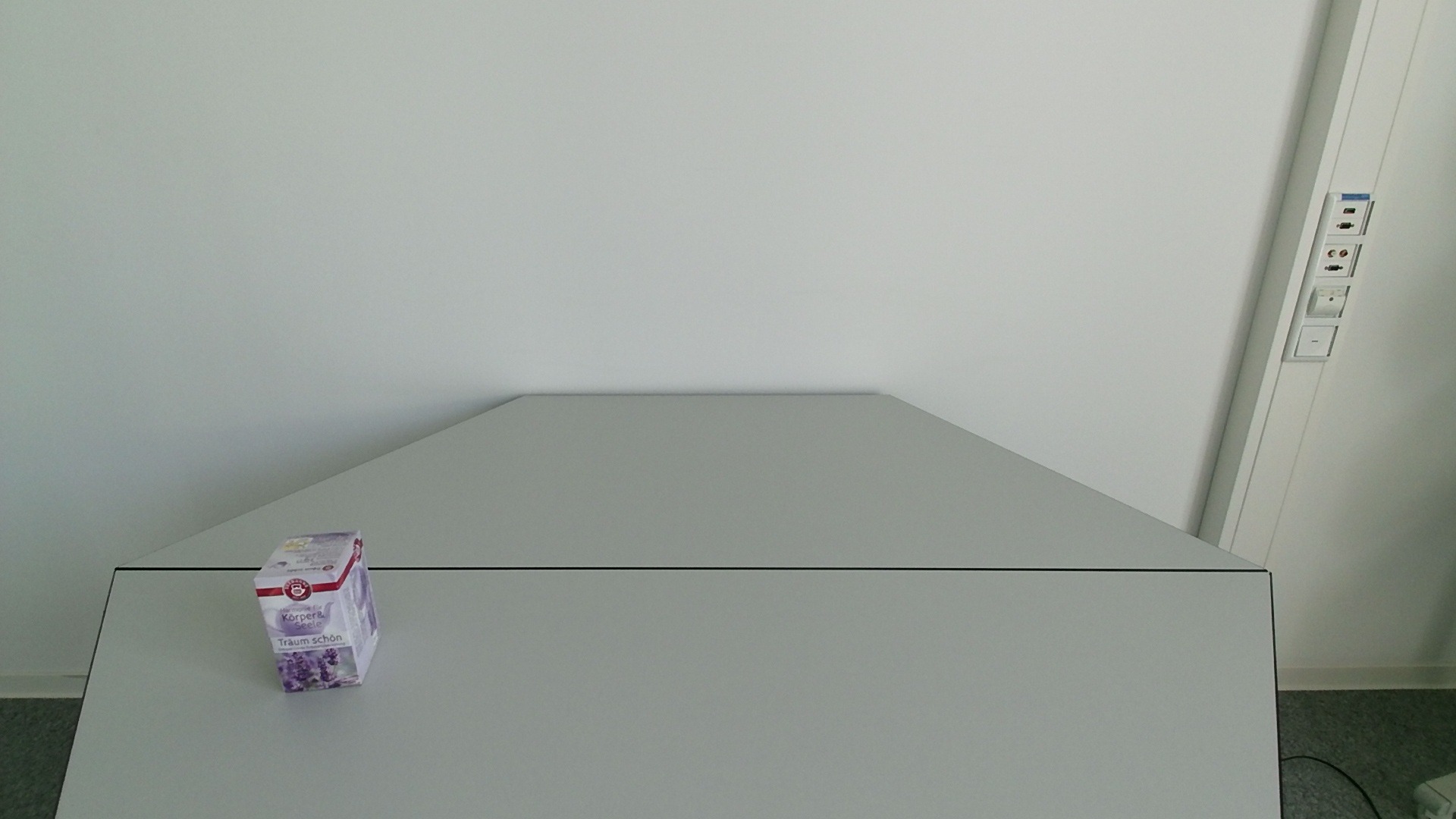}
\put (30,62) {Input}
\end{overpic}&
\begin{overpic}[width=0.135\linewidth]{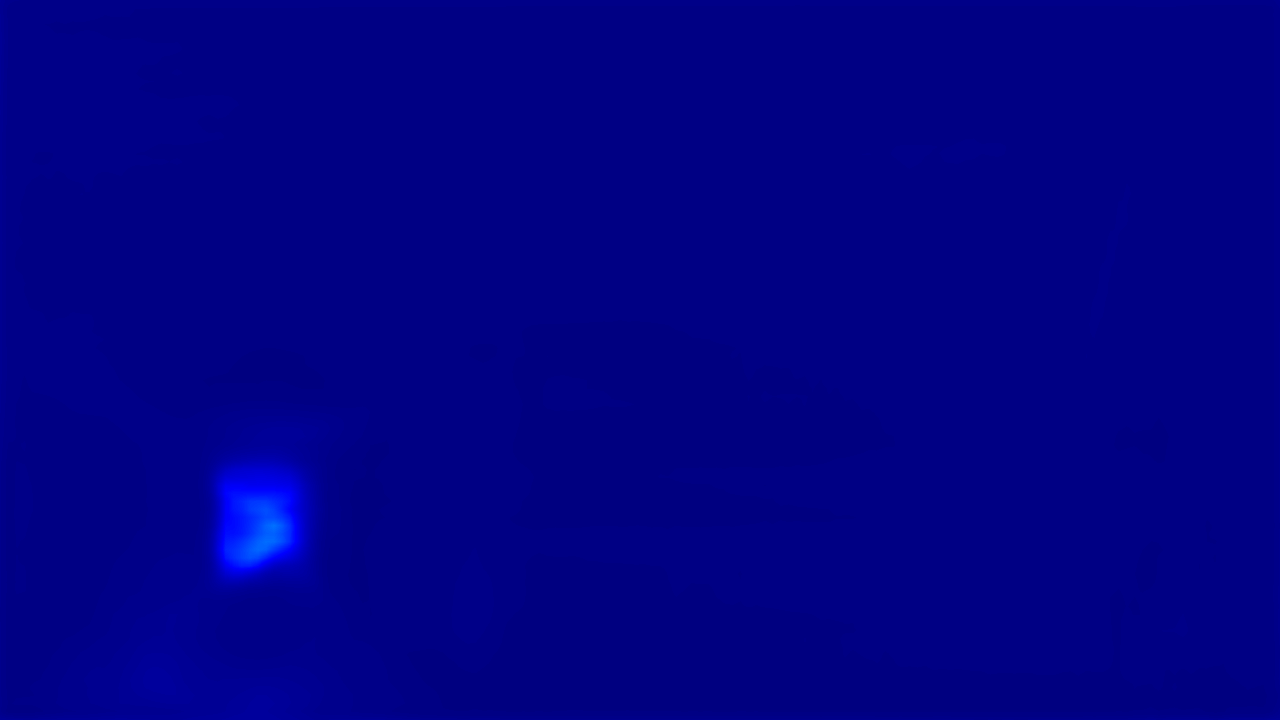}
\put (30,62) {Inside}
\end{overpic}&
\begin{overpic}[width=0.135\linewidth]{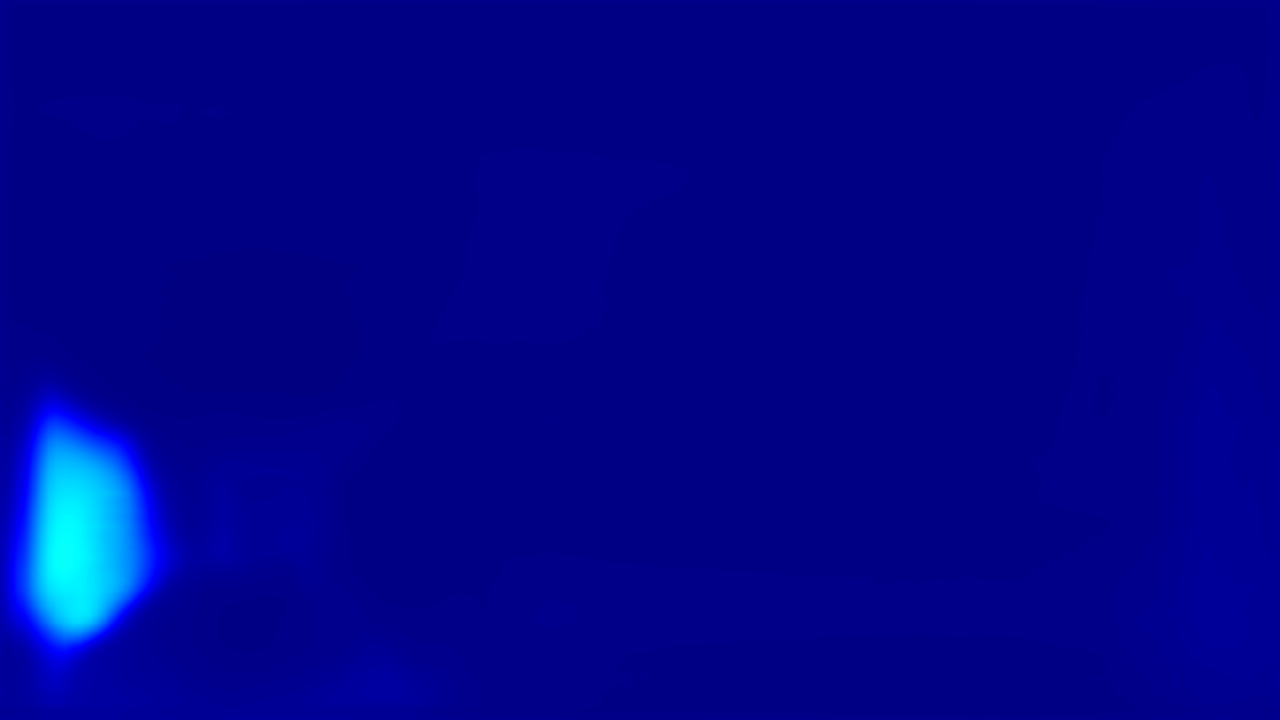}
\put (30,62) {Left}
\end{overpic}&
\begin{overpic}[width=0.135\linewidth]{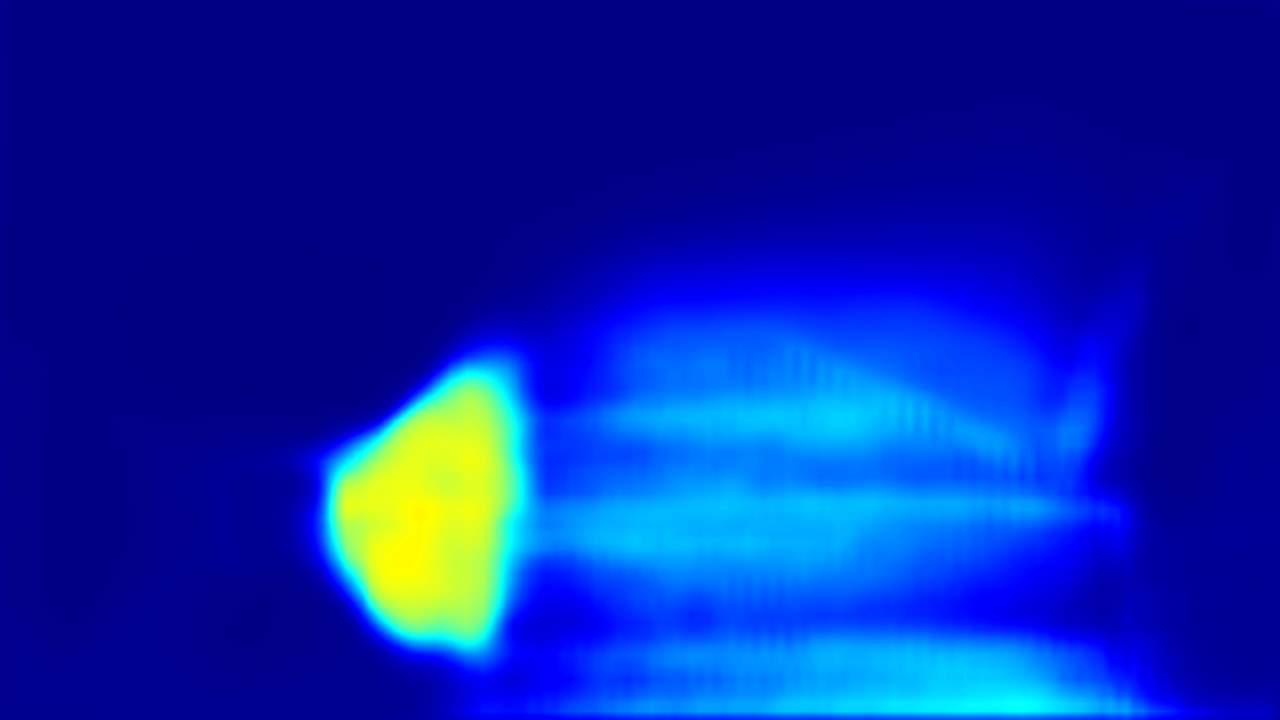}
\put (30,62) {Right}
\end{overpic}&
\begin{overpic}[width=0.135\linewidth]{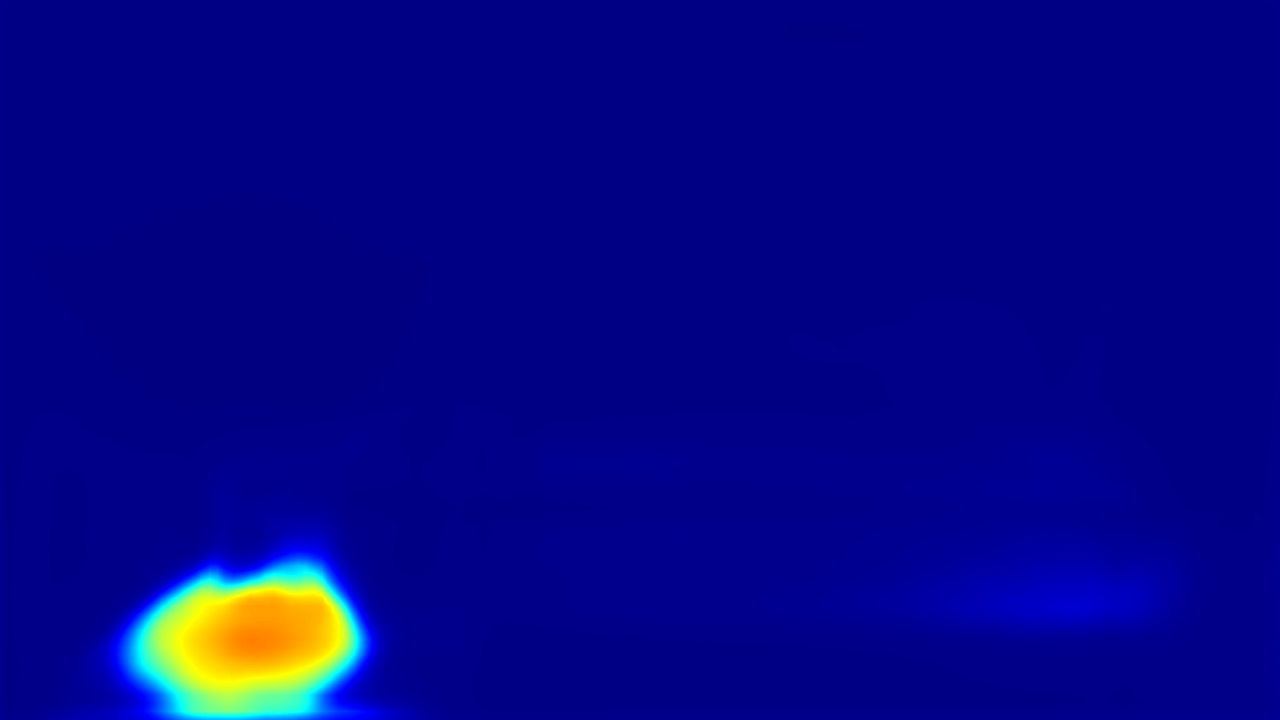}
\put (25,62) {In Front}
\end{overpic}&
\begin{overpic}[width=0.135\linewidth]{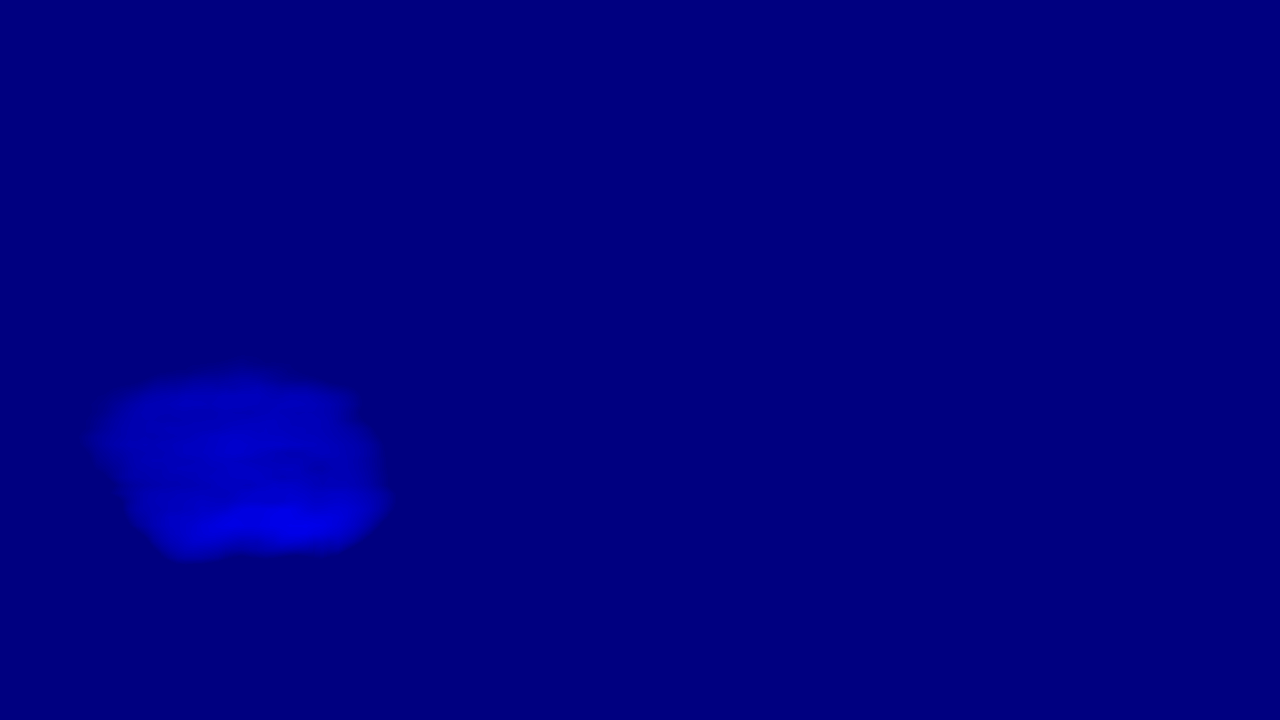}
\put (30,62) {Behind}
\end{overpic}&
\begin{overpic}[width=0.135\linewidth]{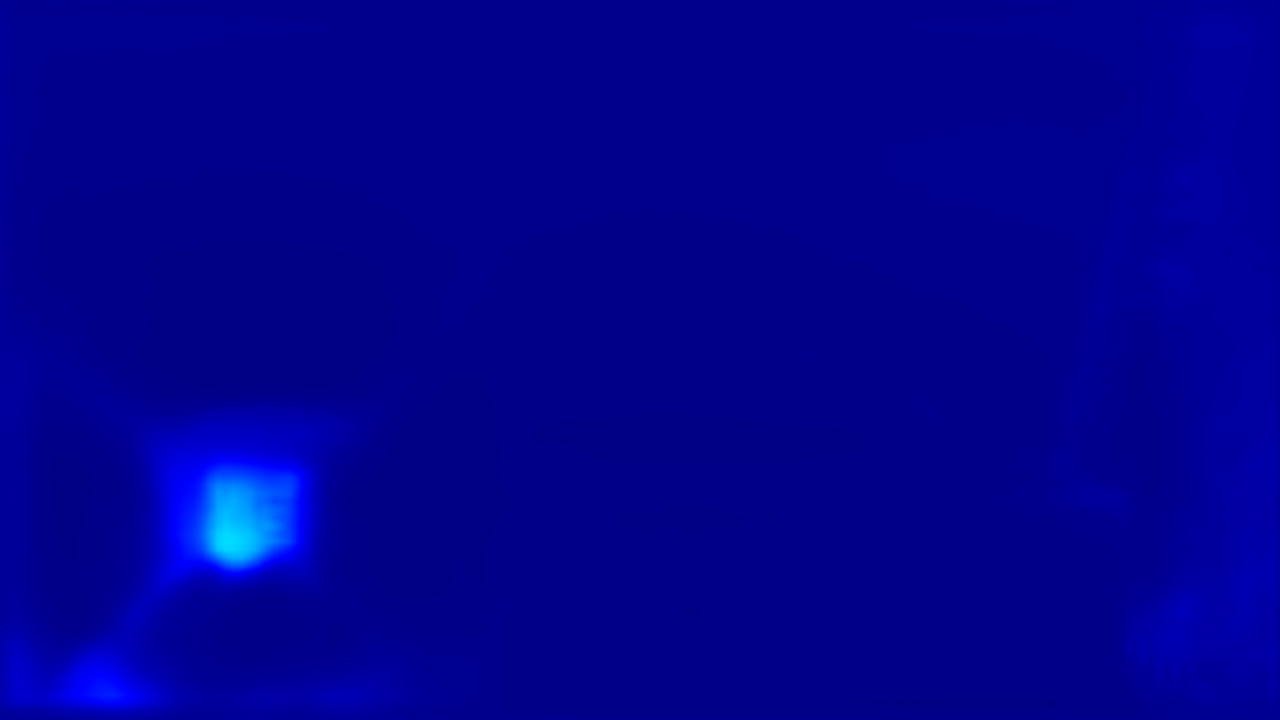}
\put (25,62) {On Top}
\end{overpic}\\
\includegraphics[width=0.135\linewidth]{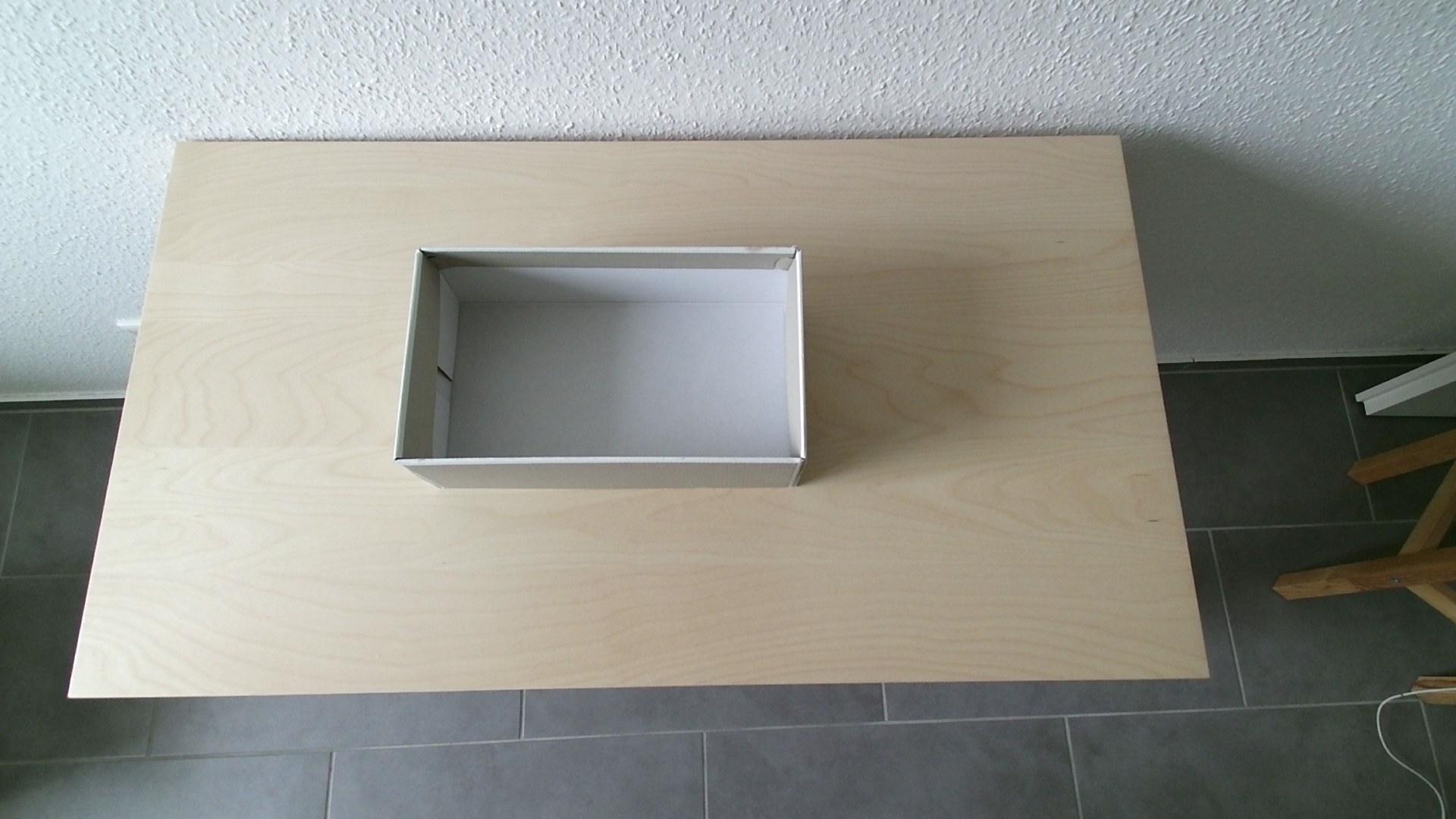}&
\includegraphics[width=0.135\linewidth]{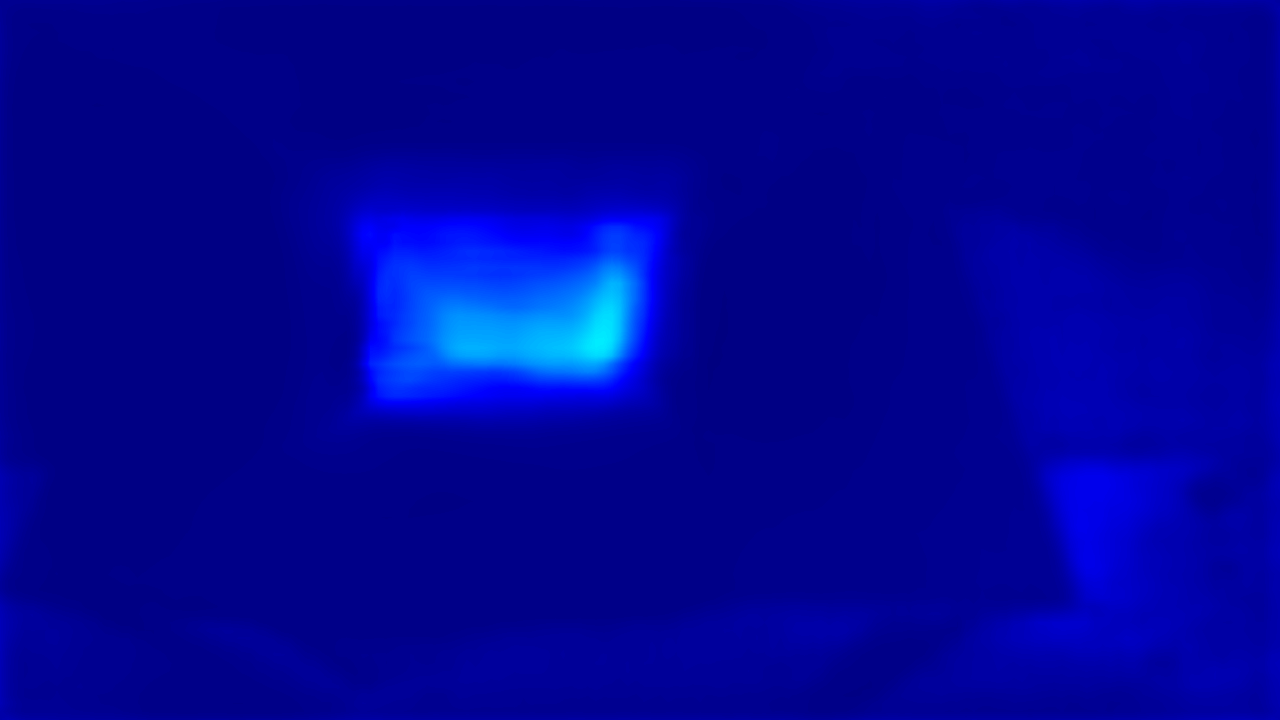}&
\includegraphics[width=0.135\linewidth]{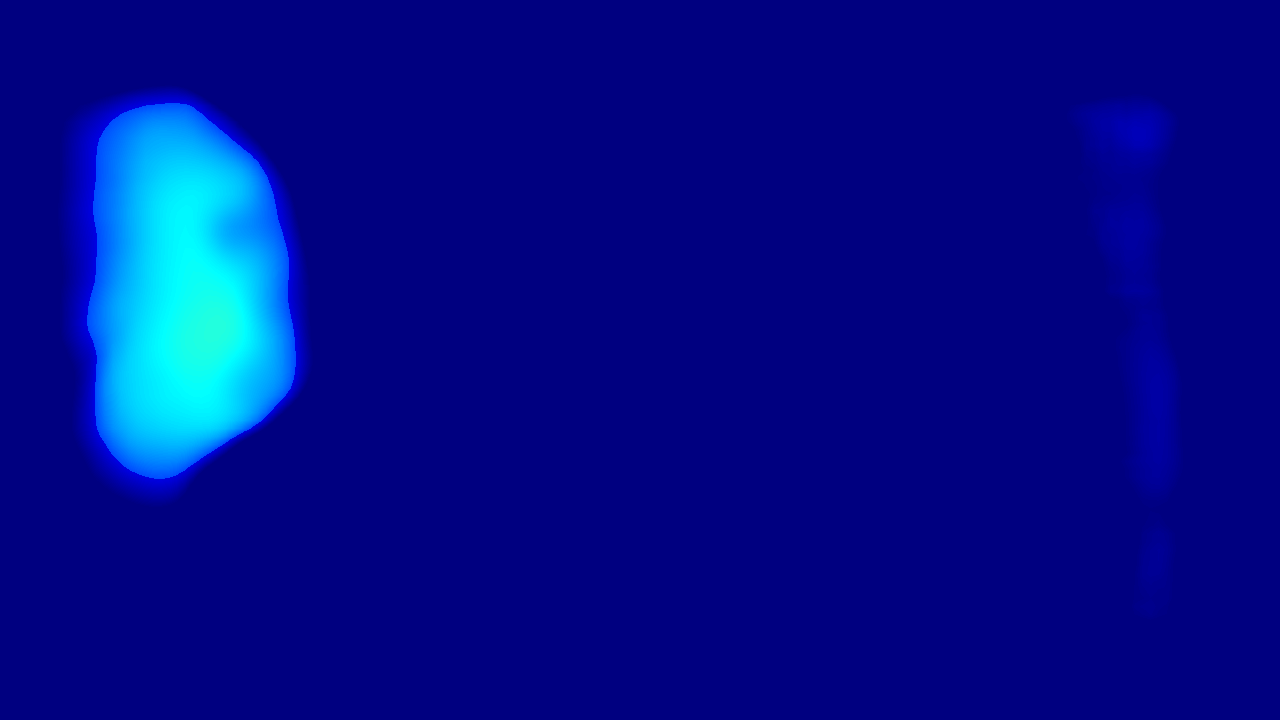}&
\includegraphics[width=0.135\linewidth]{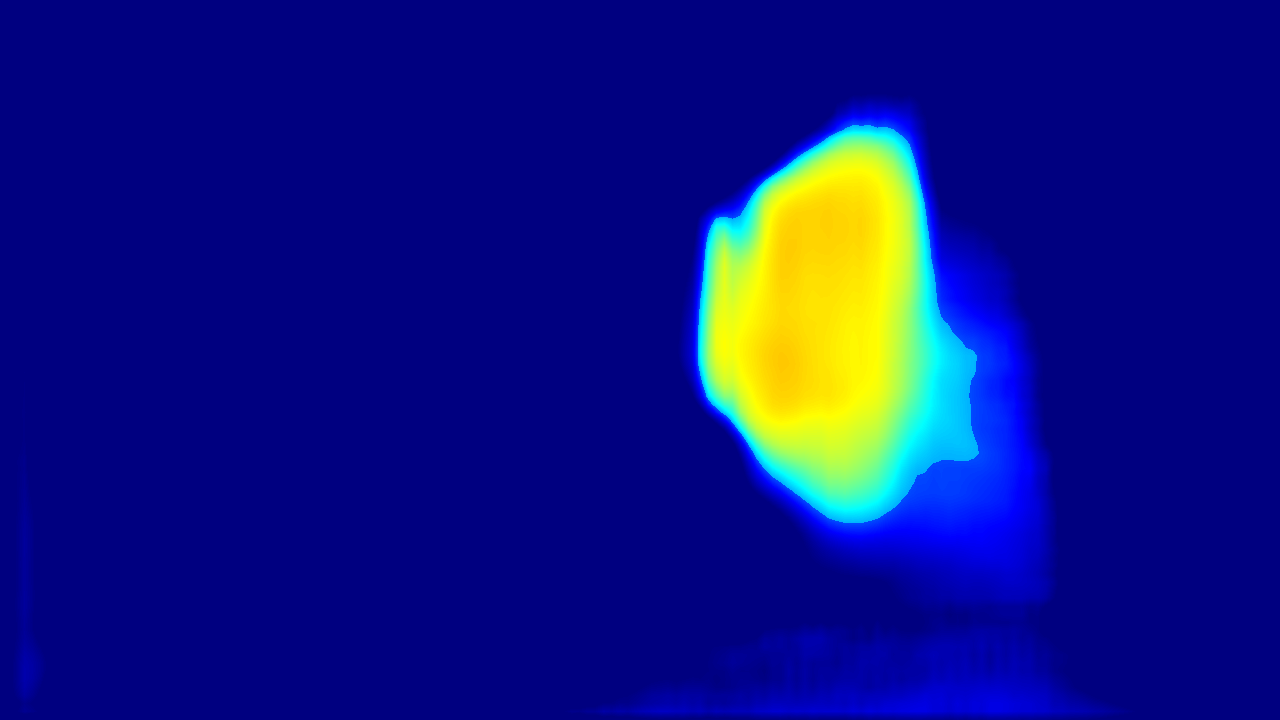}&
\includegraphics[width=0.135\linewidth]{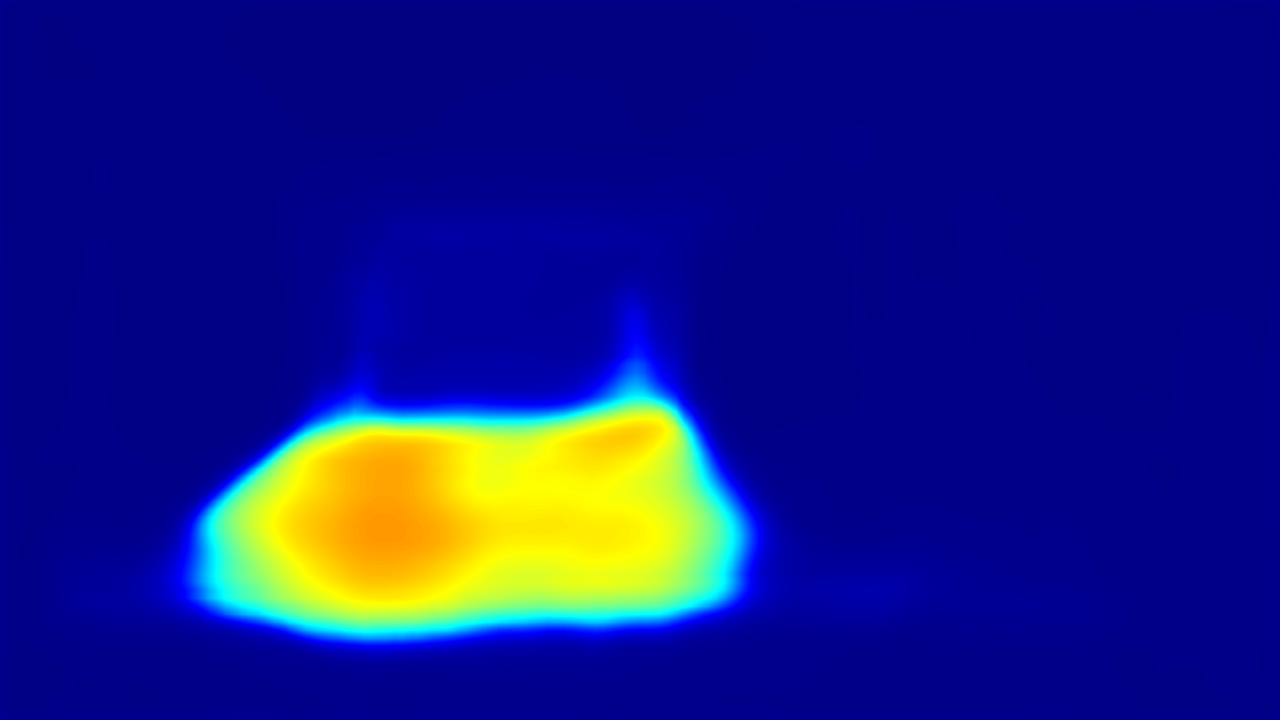}&
\includegraphics[width=0.135\linewidth]{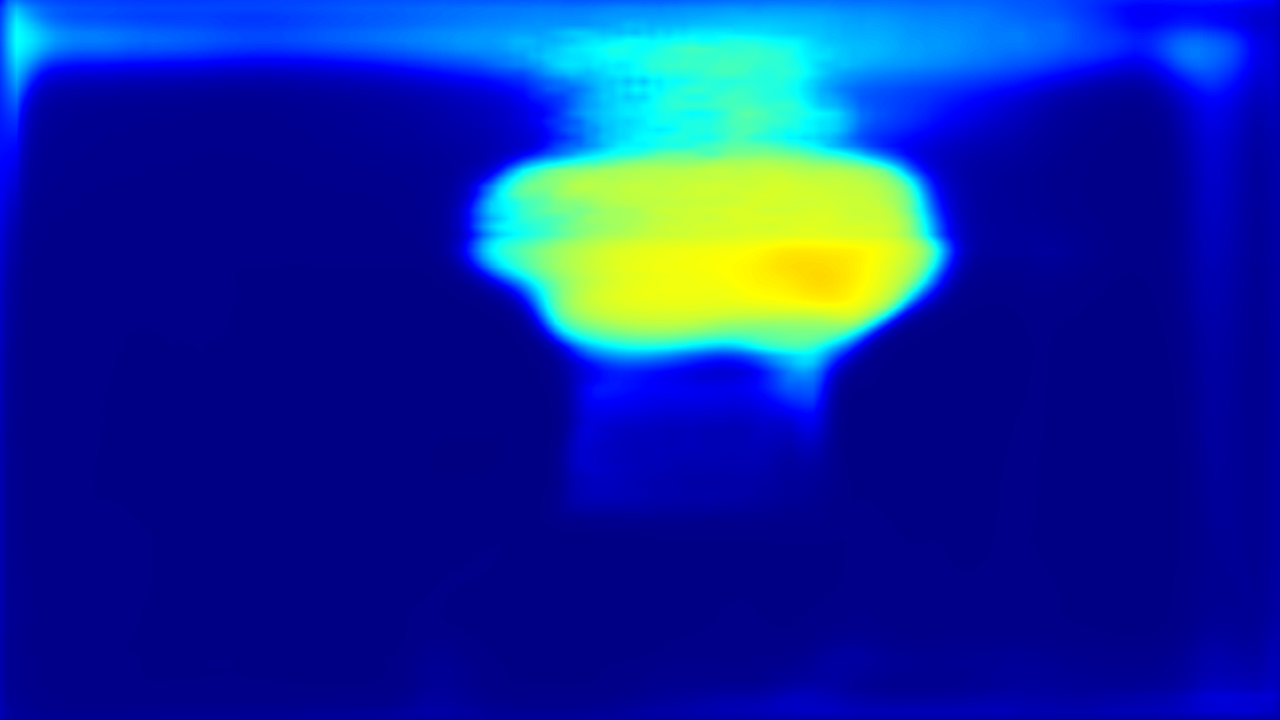}&
\includegraphics[width=0.135\linewidth]{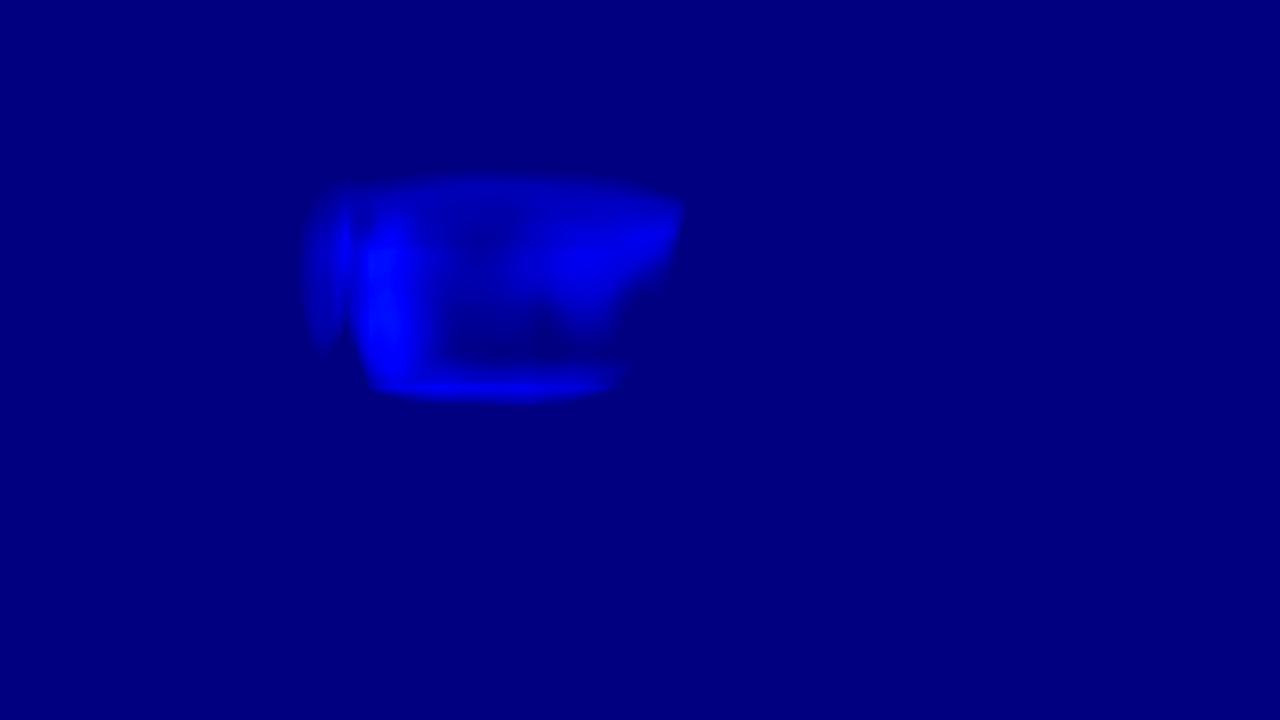}\\
\includegraphics[width=0.135\linewidth]{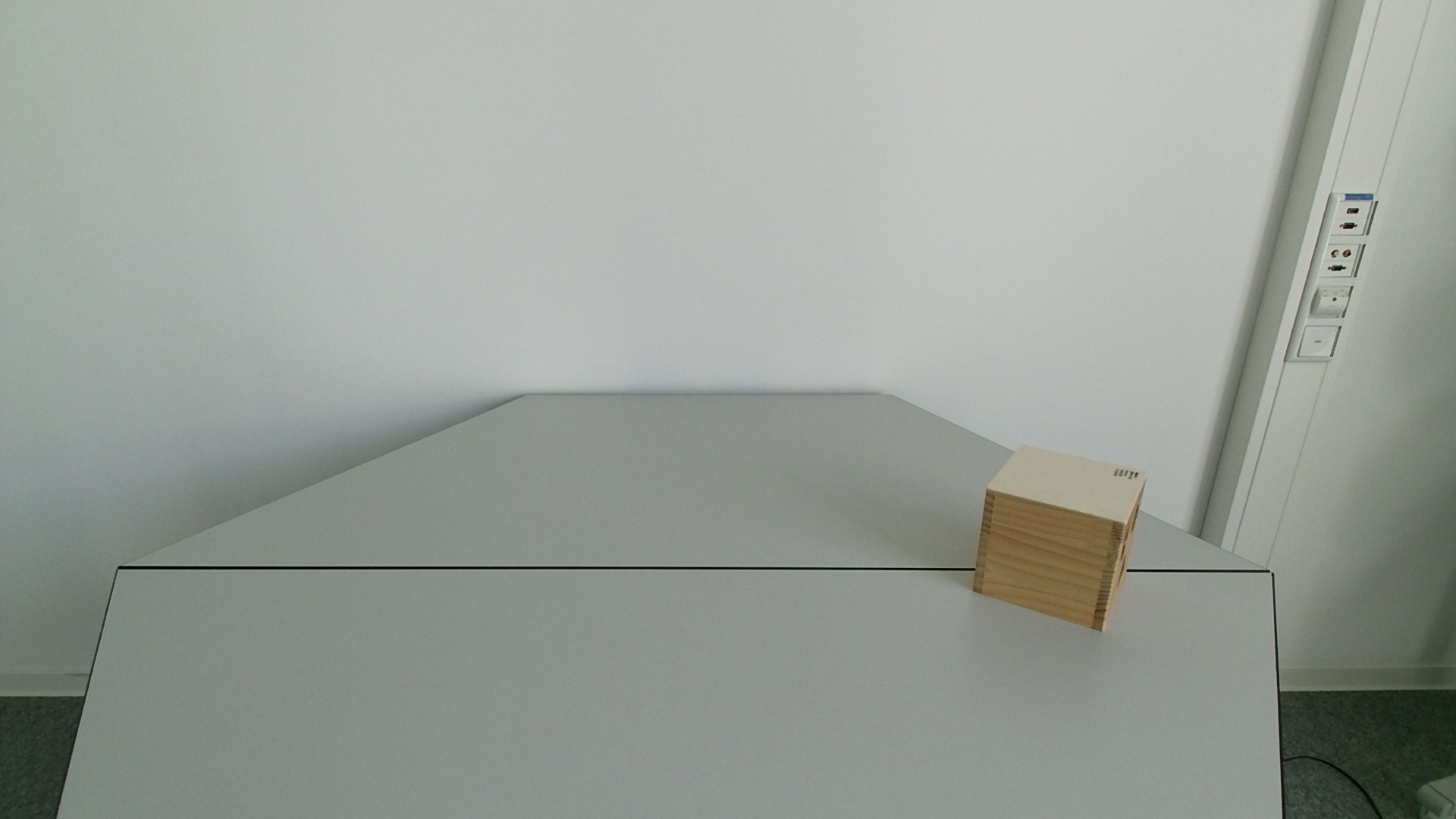}&
\includegraphics[width=0.135\linewidth]{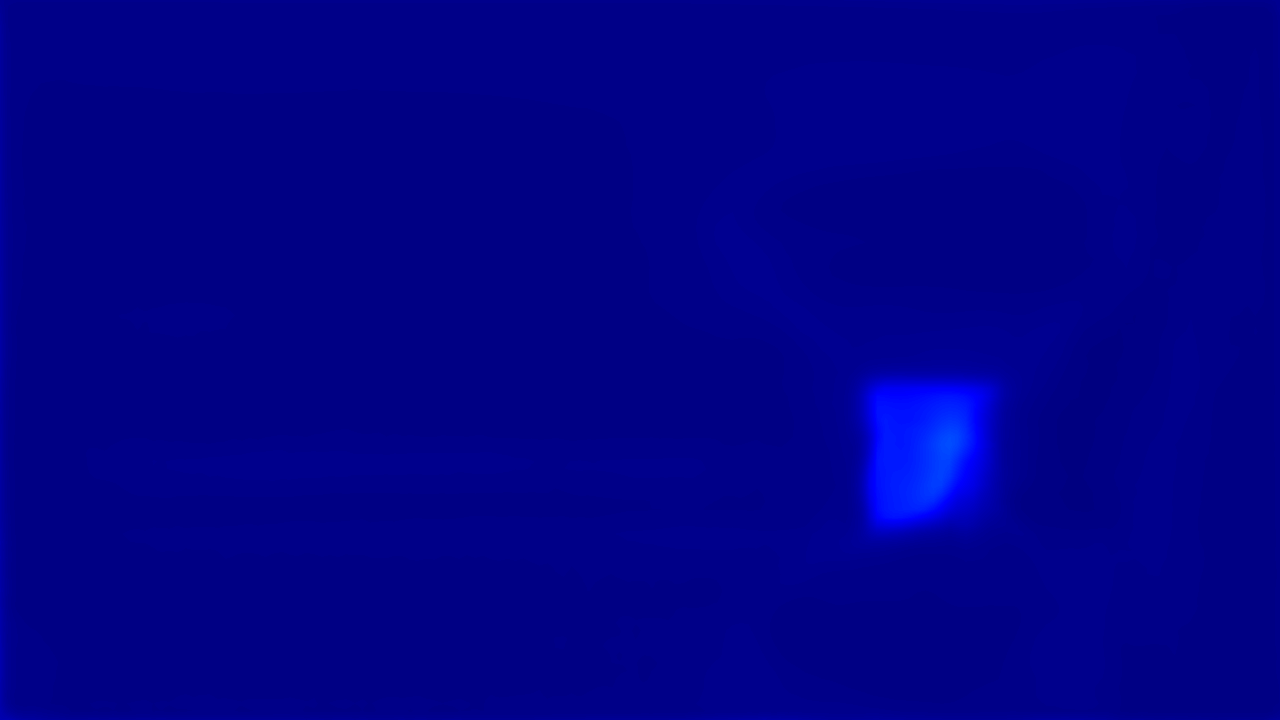}&
\includegraphics[width=0.135\linewidth]{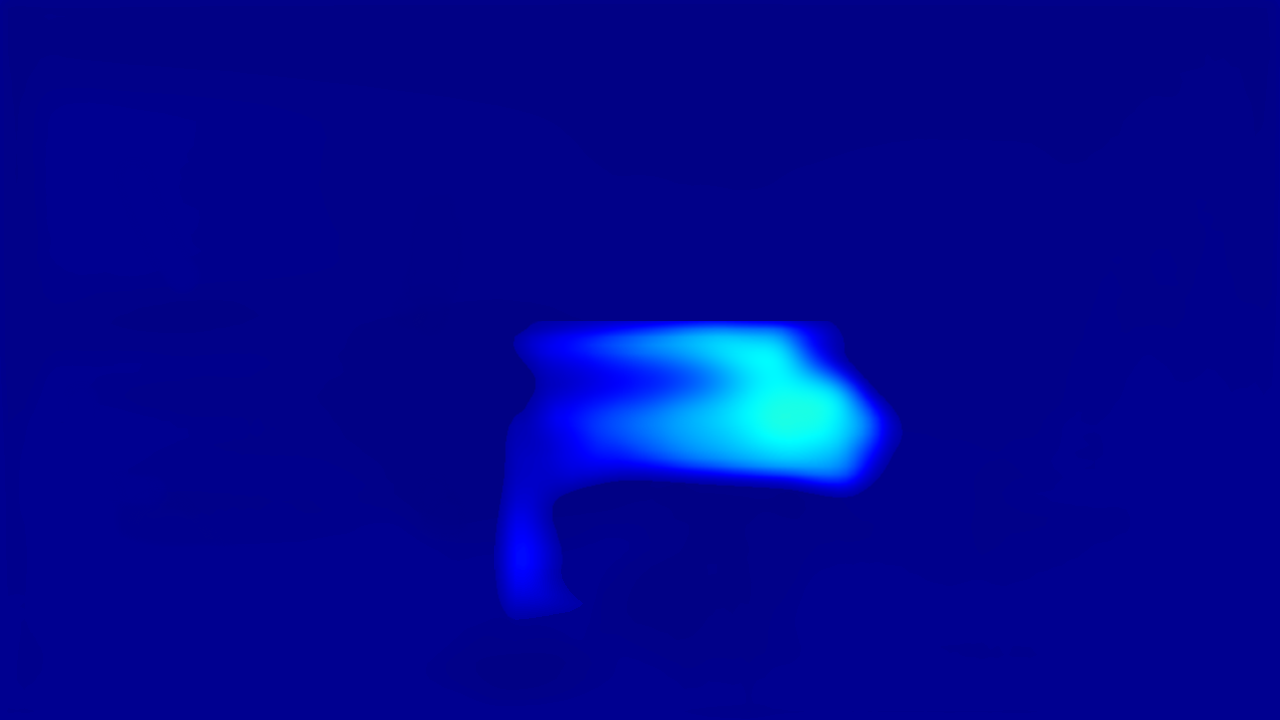}&
\includegraphics[width=0.135\linewidth]{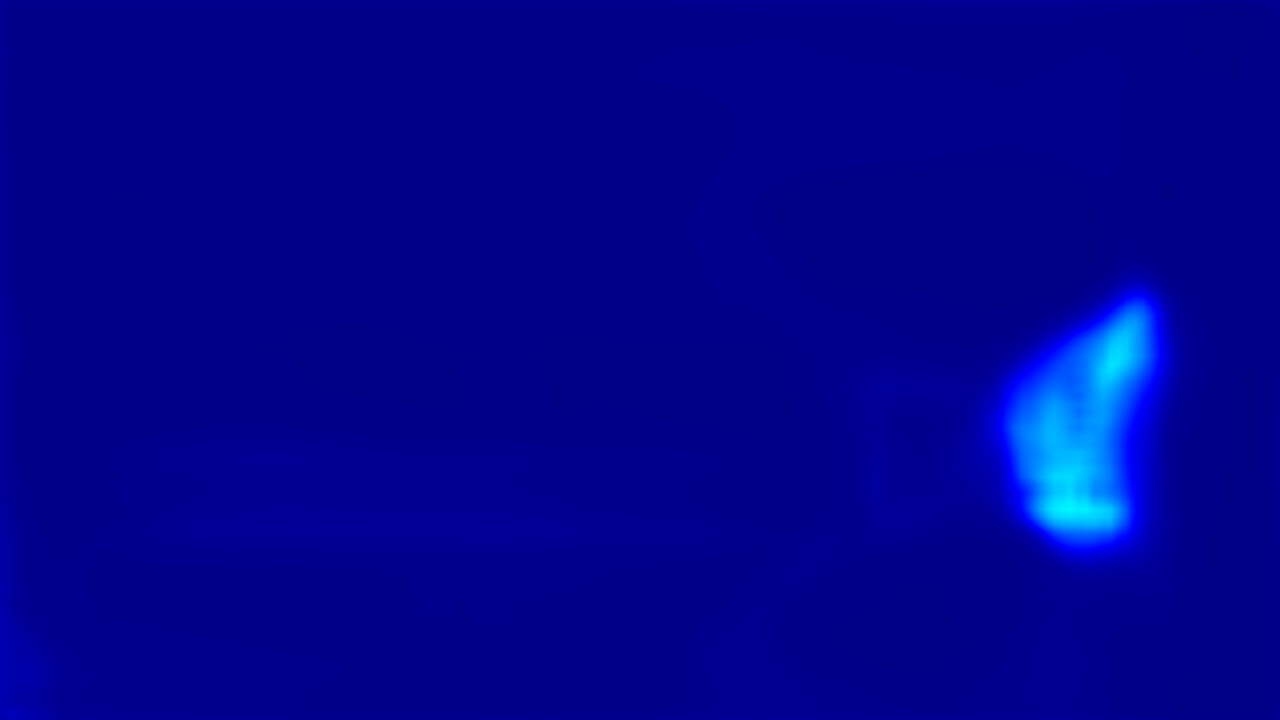}&
\includegraphics[width=0.135\linewidth]{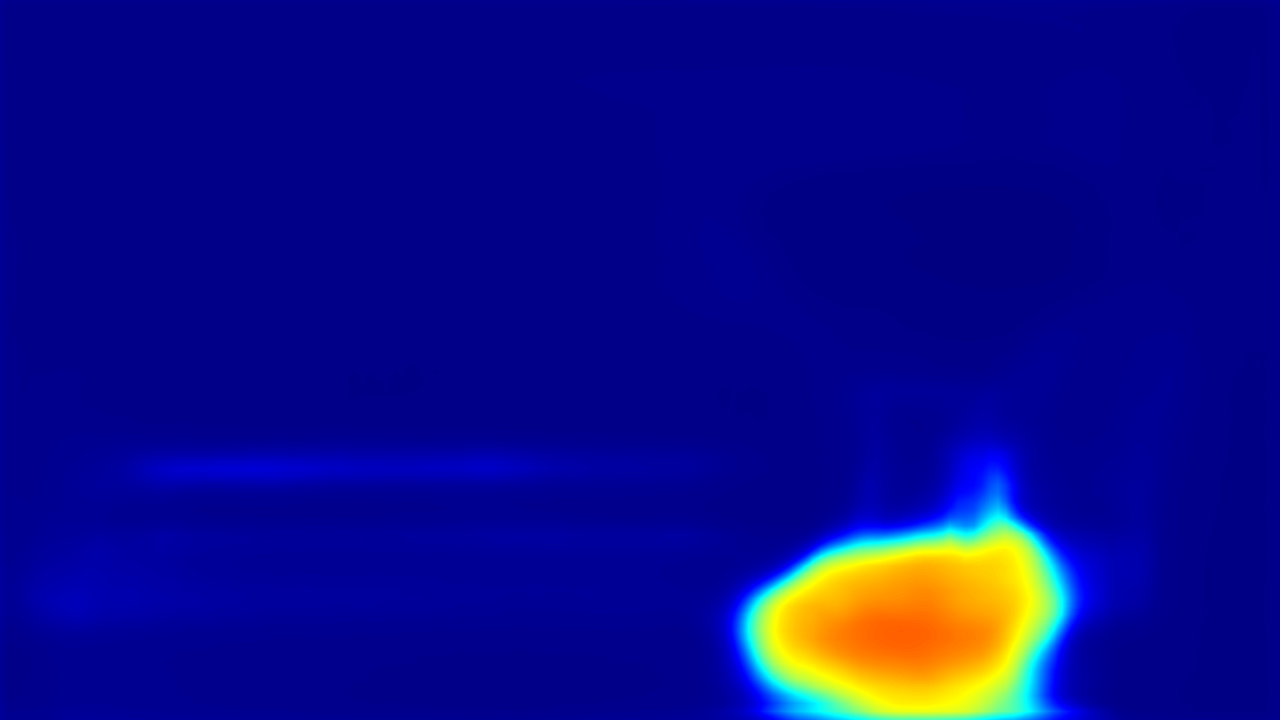}&
\includegraphics[width=0.135\linewidth]{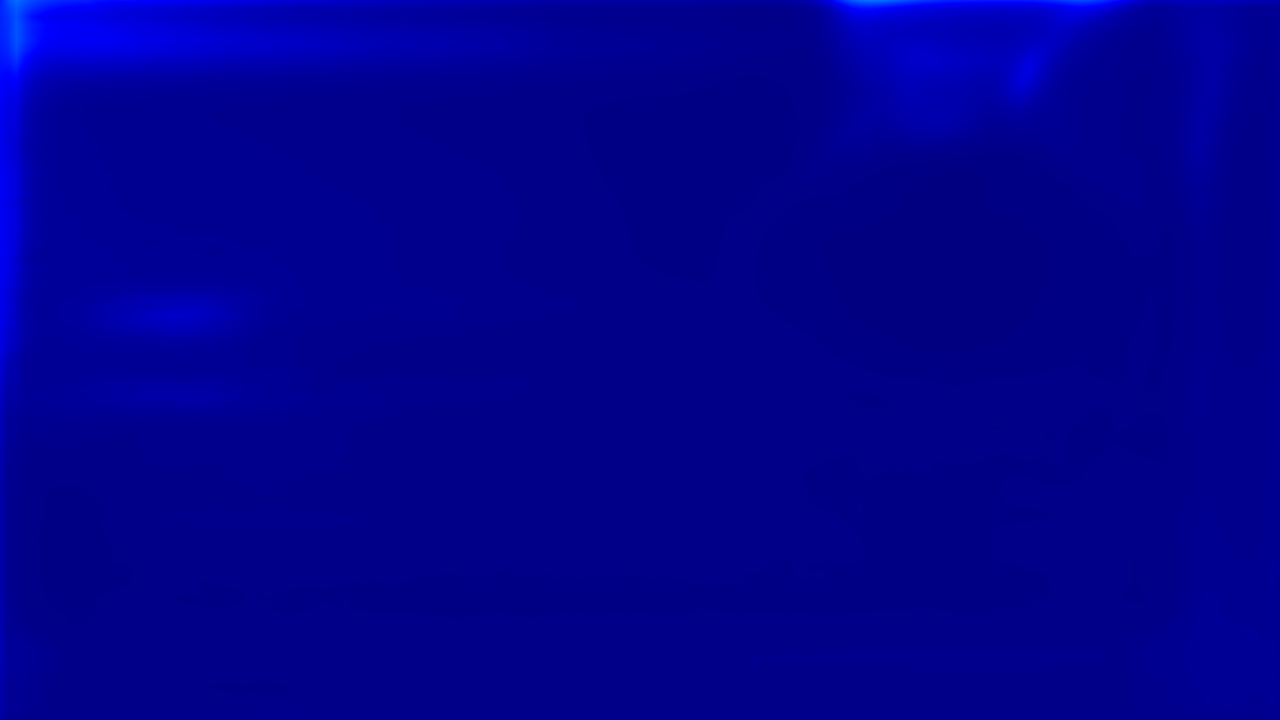}&
\includegraphics[width=0.135\linewidth]{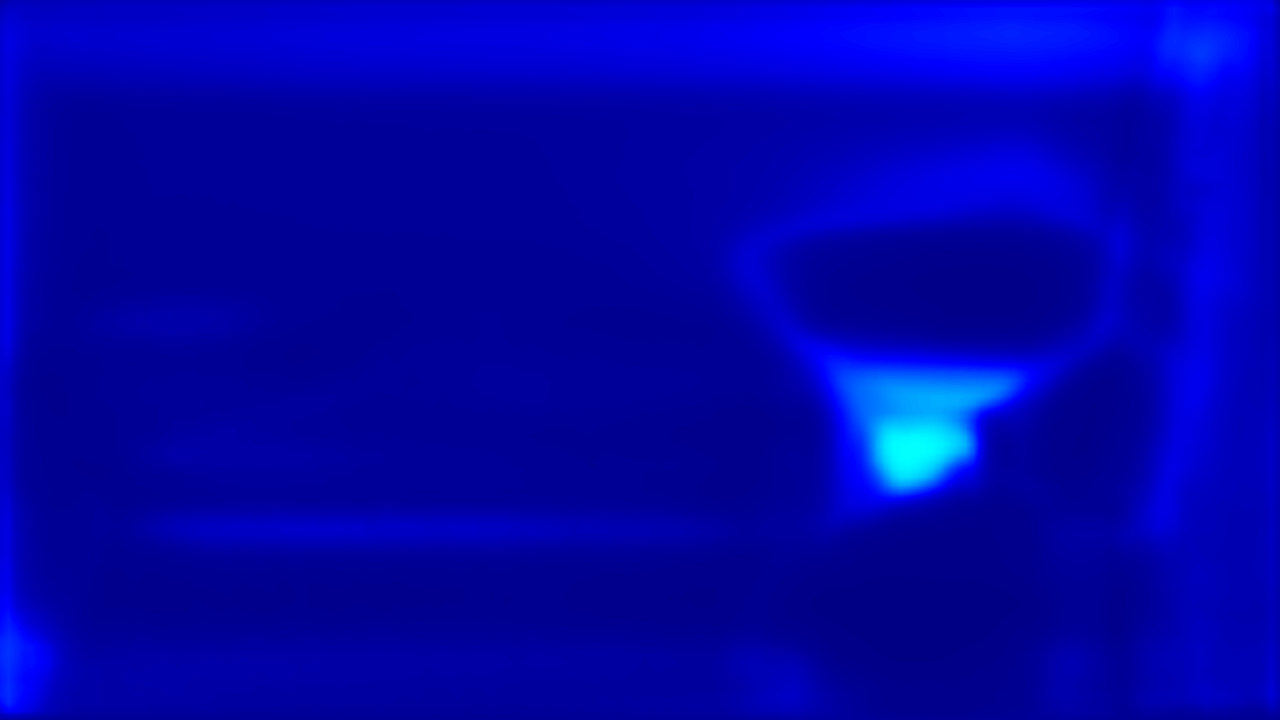}\\
\includegraphics[width=0.135\linewidth]{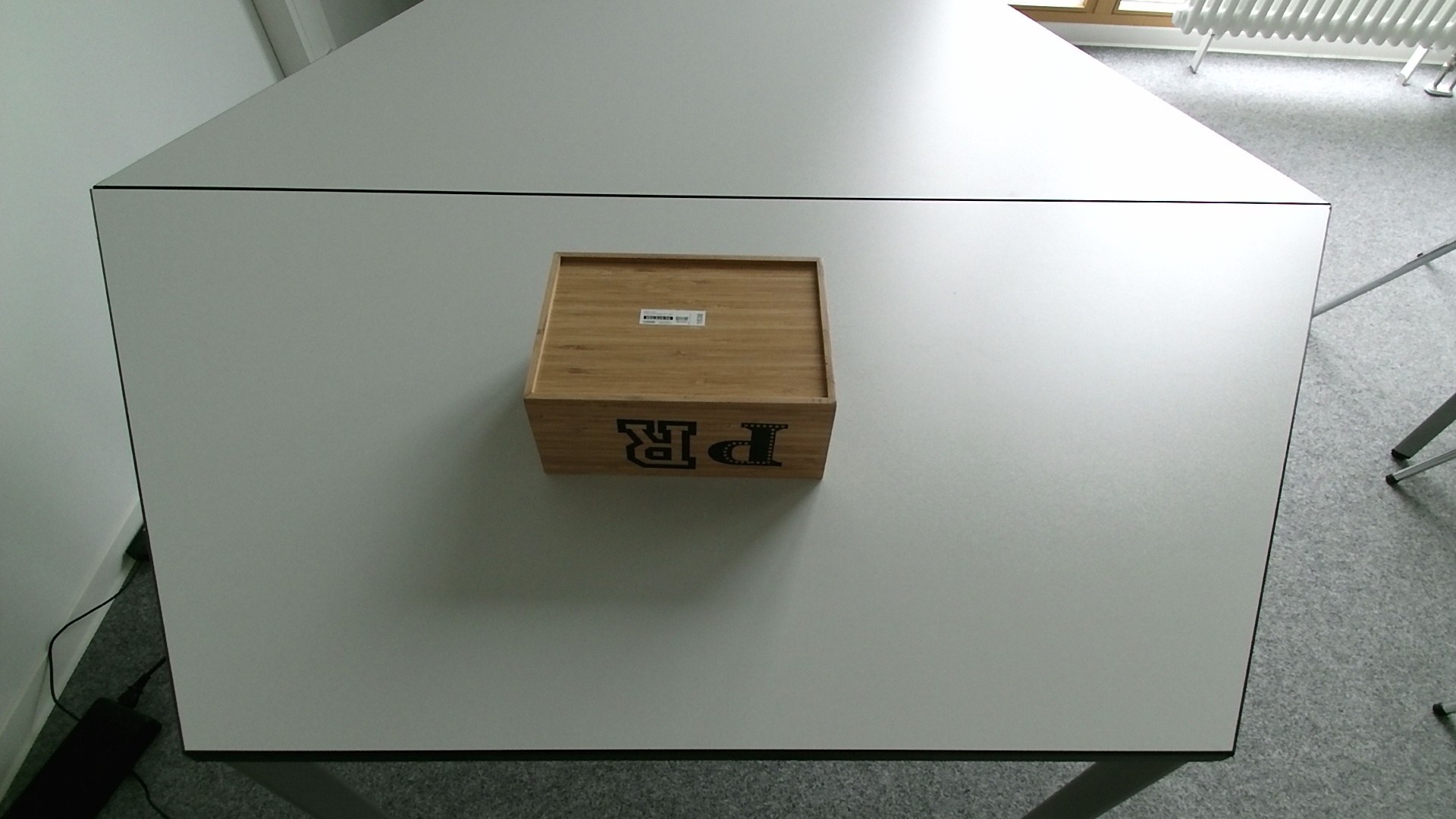}&
\includegraphics[width=0.135\linewidth]{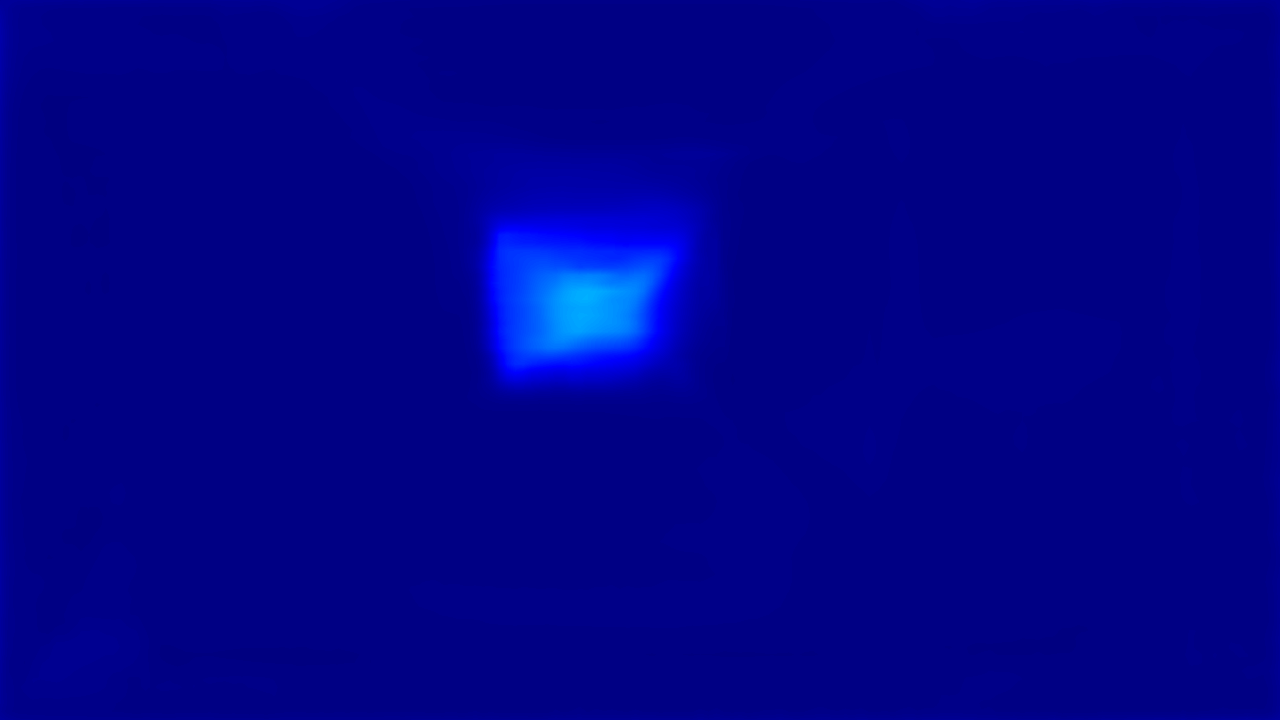}&
\includegraphics[width=0.135\linewidth]{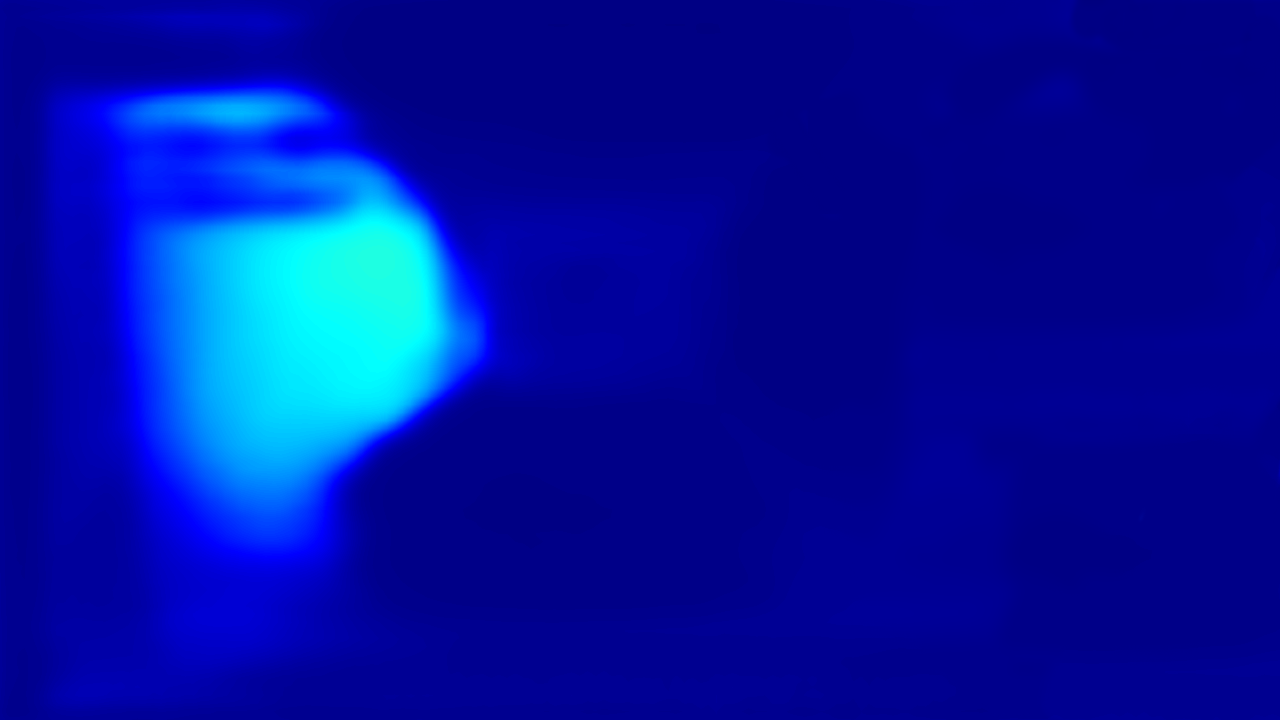}&
\includegraphics[width=0.135\linewidth]{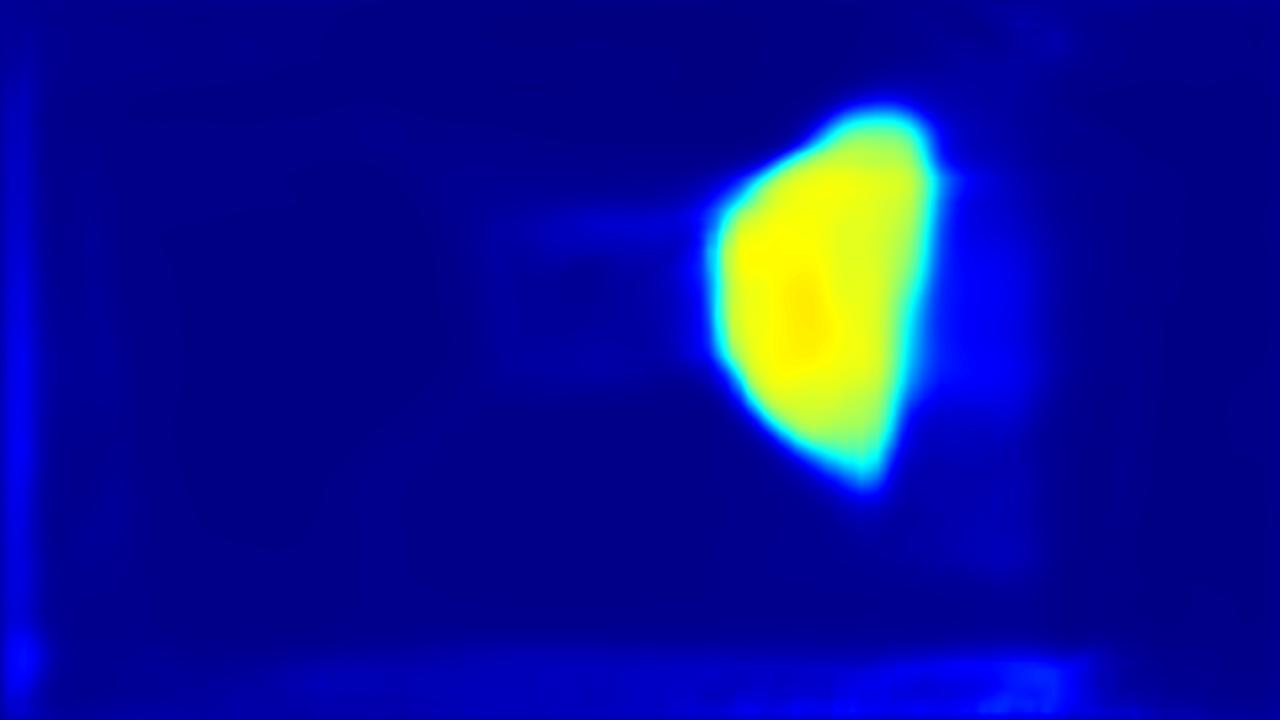}&
\includegraphics[width=0.135\linewidth]{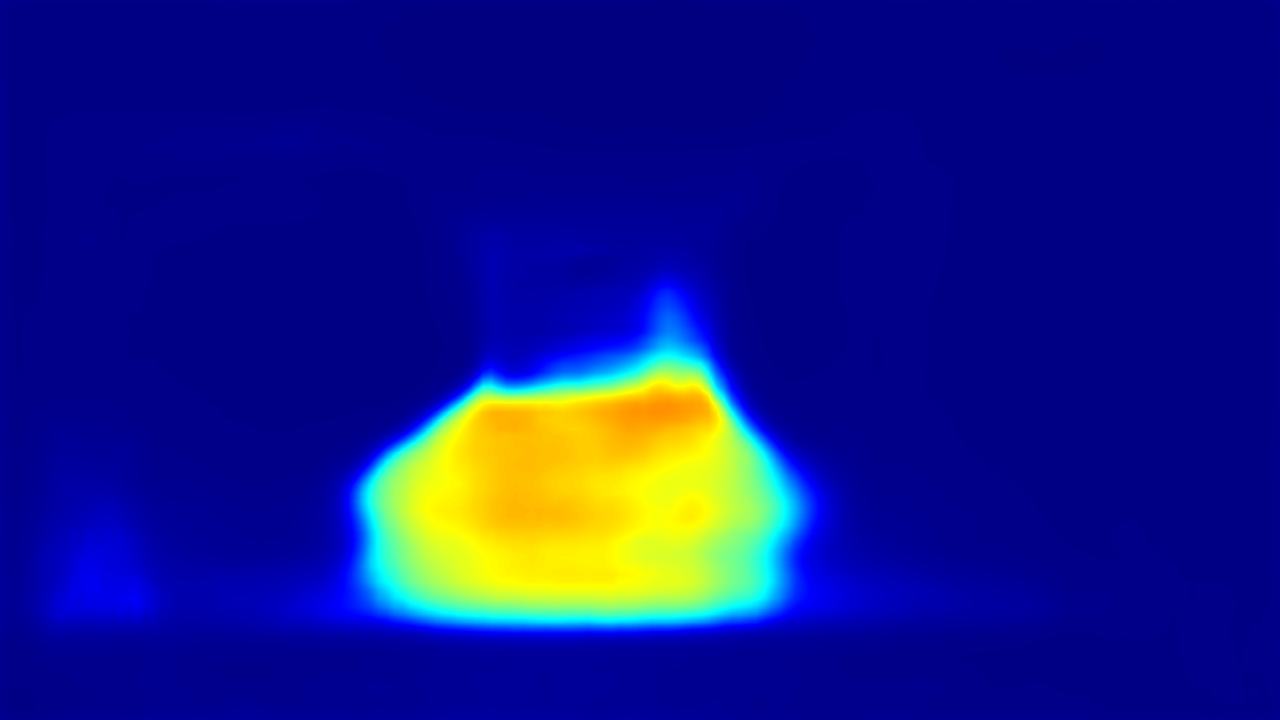}&
\includegraphics[width=0.135\linewidth]{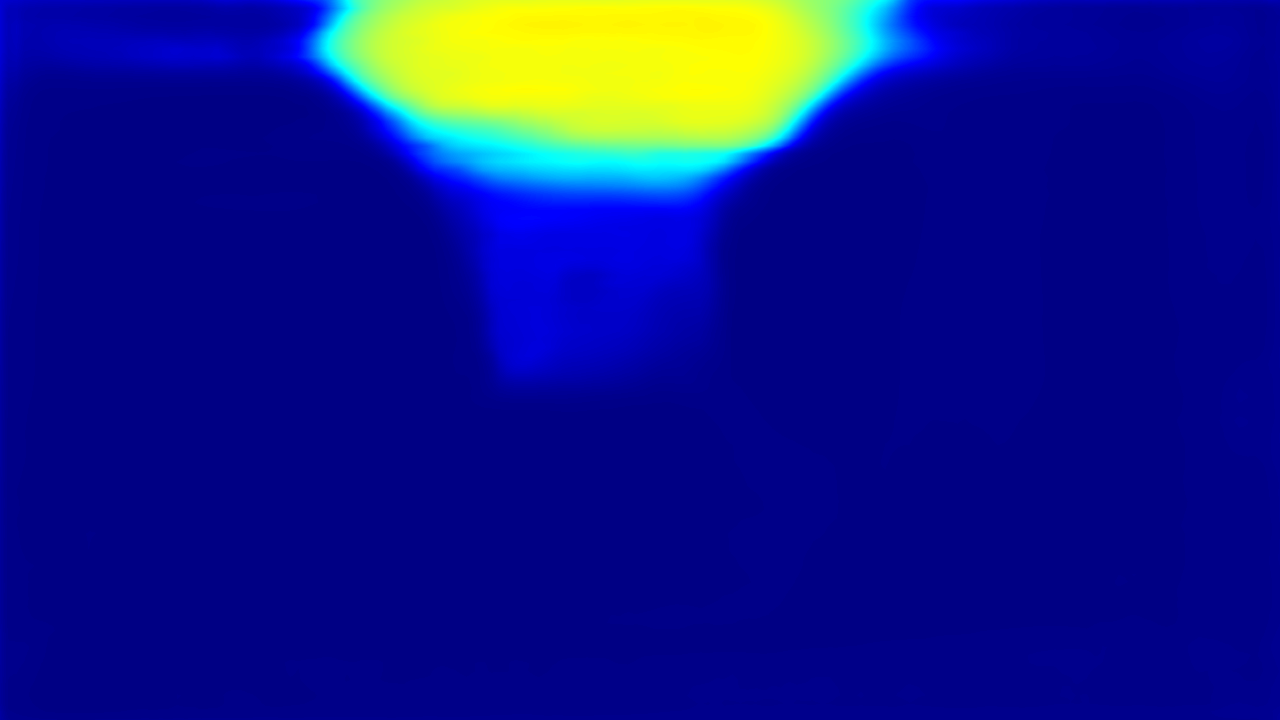}&
\includegraphics[width=0.135\linewidth]{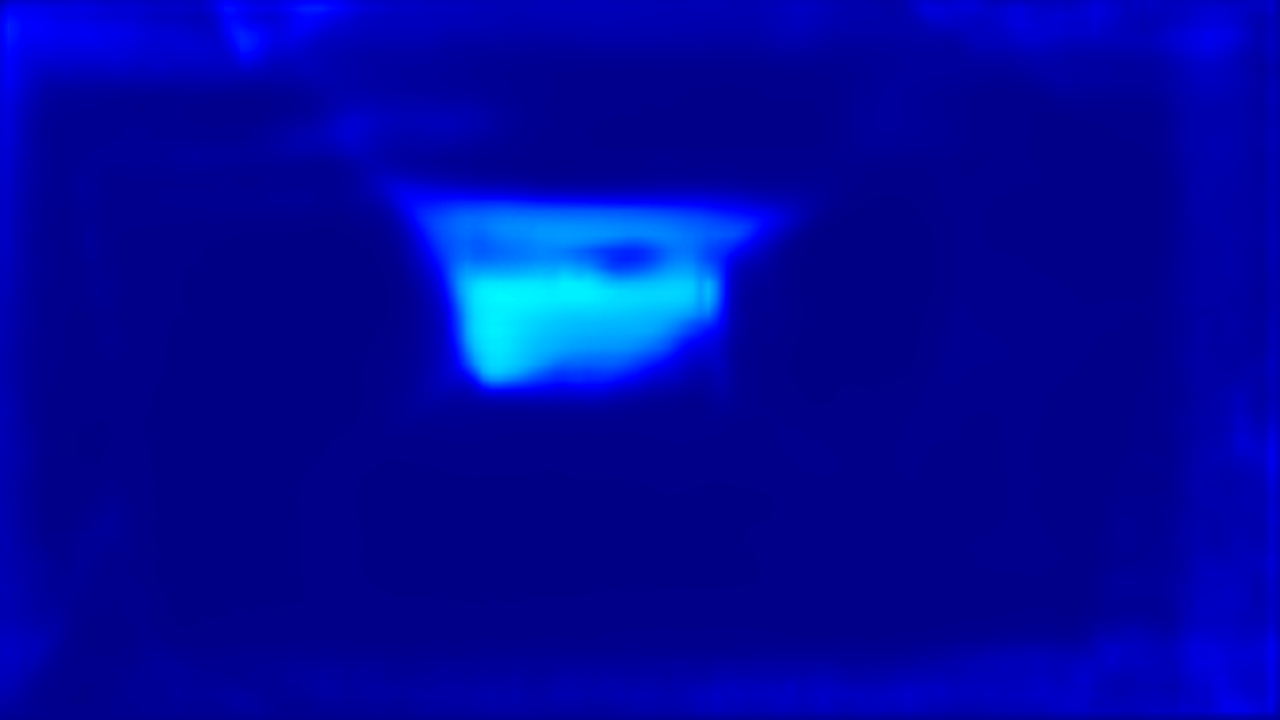}\\
\end{tabular}

\caption{Qualitative results for predicting pixelwise distributions for every spatial relation. Placing an object at a location sampled from these distributions maximizes the probability of reproducing the selected spatial relation. Our network produces meaningful distributions, despite relying solely on an auxiliary task of classifying hallucinated high-level scene representations into a set of spatial relations for supervision. }
\label{fig:pred_distributions}
\end{figure*}

\subsection{Evaluation protocol}
To compare the pixelwise distributions predicted by our method with the ground-truth pixelwise annotations provided by the participants, we report several metrics. Inspired by metrics from object detection and segmentation, we threshold the distributions at different ranges and compute their mean intersection over union (IoU). Additionally, we are interested in comparing the modes between the distributions. First, we compute the maximum mode from each distribution and report the euclidean pixel distance between them (Mode).  As the distance between the modes does not model the tails of the distributions, we also calculate the distance between the centroid pixels of the predicted and ground-truth distributions (Centroid). Due to the non-parametric nature of the distributions, we perform a Kruskal-Wallis (KW) test by analyzing if $100$ points sampled from the ground-truth distribution and another $100$ points sampled  from the predicted distribution originated from the same distribution, with a significance of $p<0.05$.
Finally, we measure the similarity of the probability distributions with the Kullback–Leibler (KL) and Jensen–Shannon (JS) divergences. 

\subsection{Quantitative Results} 
First, we analyze the performance of the auxiliary RelNet network to model spatial relations, as we rely on it to get the learning signal for Spatial-RelNet. We evaluate the performance of RelNet on a test split containing 975 pairwise relations and report an average accuracy of $97\%$ over all relations, as shown in Table~\ref{tab:results_relnet}. We compare its performance against a model that was trained only on binary masks of the objects to analyze the importance of using the image context to model the relations. This model achieves an accuracy of $84.4\%$ and we find that the image context is specially important to disambiguate the relations  \texttt{on top} and \texttt{inside}. We also train an intermediate model, which takes as input the image and binary object masks to model the relations, and achieves an  accuracy of $94.3\%$. Our final model shows the best performance by incorporating the use of the Gaussian distance transforms for the attention masks.

\begin{table}[h]
 \setlength{\tabcolsep}{3.5pt}
  \begin{tabular}{l | c | c c c c c c}
  Model              & Mean  & Inside & Left & Right & In Front & Behind & On Top\\
  \hline
  \hline
  Masks only         & 84.4  & 60.8   & 99.3  & 93.2 & 99.3     & 98.1   & 56.6\\
  Image + Masks     & 94.3  & 81.3   & 99.3  & 100 & 98.7     & 97.5   & 88.5\\
  Full model         & 97    & 93.1   & 98.7   & 100 & 100     & 98.7   & 91.5\\

  \end{tabular}
  \caption{Quantitative comparison of RelNet with its variants. Adding the image context helps disambiguating the relations  \texttt{on top} and \texttt{inside}.}
  \label{tab:results_relnet}
\end{table} 
Next, we quantitatively evaluate the capability of a baseline  model in which we naively ``paste'' objects masks in the RGB images to predict pixelwise distributions for spatial relations. We report mean Ious of 0.44, 0.4, 0.3 for the thresholds of 0.25, 0.5 and 0.75 respectively. This shows that the artifacts created by ``pasting'' object masks in RGB images lead to noticeably different features and to the training erroneously focusing on these discrepancies. In comparison, our Spatial-RelNet achieves  mean IoUs of 0.63, 0.6 and 0.44, as shown in Table~\ref{tab:results}, by classifying hallucinated scene representations, alleviating this problem. Moreover, for Spatial-RelNet the  mean distance between the modes of the distributions lies at 67.2 pixels, corresponding approximately to 5.74cm. 
As this metric depends on the image resolution and the distance of the camera to the objects, for each image in the test set, we sample uniformly $100$ pixels and compute their average distance to the mode of the ground-truth distribution. Thus, the mean distance between a random pixel and the ground-truth mode is 504.5 pixels.
To model the tails of the distributions, we also calculate the distance between the centroid pixels of the predicted and ground-truth distributions  and we report a  mean distance of 113.5 pixels, which corresponds approximately to 9.88 cm.
\begin{table}[h]
 \setlength{\tabcolsep}{4.5pt}
  \begin{tabular}{l | c | c c c c c c}
  Metric    & Mean  & Inside & Left & Right & In Front & Behind & On Top\\
  \hline
  \hline
  \ioua     & 0.63  & 0.66   & 0.69  & 0.65 & 0.64  & 0.51  & 0.65\\
  \ioub     & 0.6  & 0.62   & 0.6  & 0.62 & 0.57   &  0.57  & 0.62\\
  \iouc     & 0.44  & 0.47   & 0.41  & 0.46 & 0.39  & 0.48 & 0.5\\
  Mode      & 67.2  & 43.5   & 71  & 59.2 & 86.8    & 115.3  & 90.5\\
  Centroid  & 113.5 & 116.7  & 241.1 & 85.1  & 70.9     & 51.4   & 163.4\\
  KL        & 3.78  & 5.35   & 4.47  & 3.5  & 1.62     & 3.9    & 5.7\\
  JS        & 0.46  & 0.54   & 0.48  & 0.45 & 0.34     & 0.5    & 0.57\\
  KW        & 0.55  & 0.63    & 0.5   & 0.51 & 0.51     & 0.43   & 0.73\\
  \end{tabular}
  \caption{Quantitative comparison of the predicted pixelwise distributions with ground-truth annotations for a range of metrics. Our methods yields good results though relying on a weaker form of supervision.}
  \label{tab:results}
\end{table} 
We measure the similarity of the distributions with the Kullback–Leibler and Jensen–Shannon divergences, were we report mean values of  3.78 and 0.46 respectively. Finally, we found in 0.55\% of the cases the samples drawn from the predicted and ground-truth distribution to originate from the same distribution according to the  Kruskal-Wallis test, with a significance of $p<0.05$.

We show qualitative results in Figure~\ref{fig:pred_distributions}. In this  challenging setting, the network learns to produce meaningful  distributions, from which one can sample object placement locations to reproduce a spatial relation.

\subsection{Human-Robot Object Placement Experiment}
We also evaluate the performance of our approach in a realistic human-robot collaboration context. We exemplify the ability of our approach to reason about the best way to place objects by asking a group of participants to provide relational natural language instructions to a PR2 service robot in a tabletop scene.
\begin{figure}[t]
\centering
\includegraphics[width=0.79\linewidth]{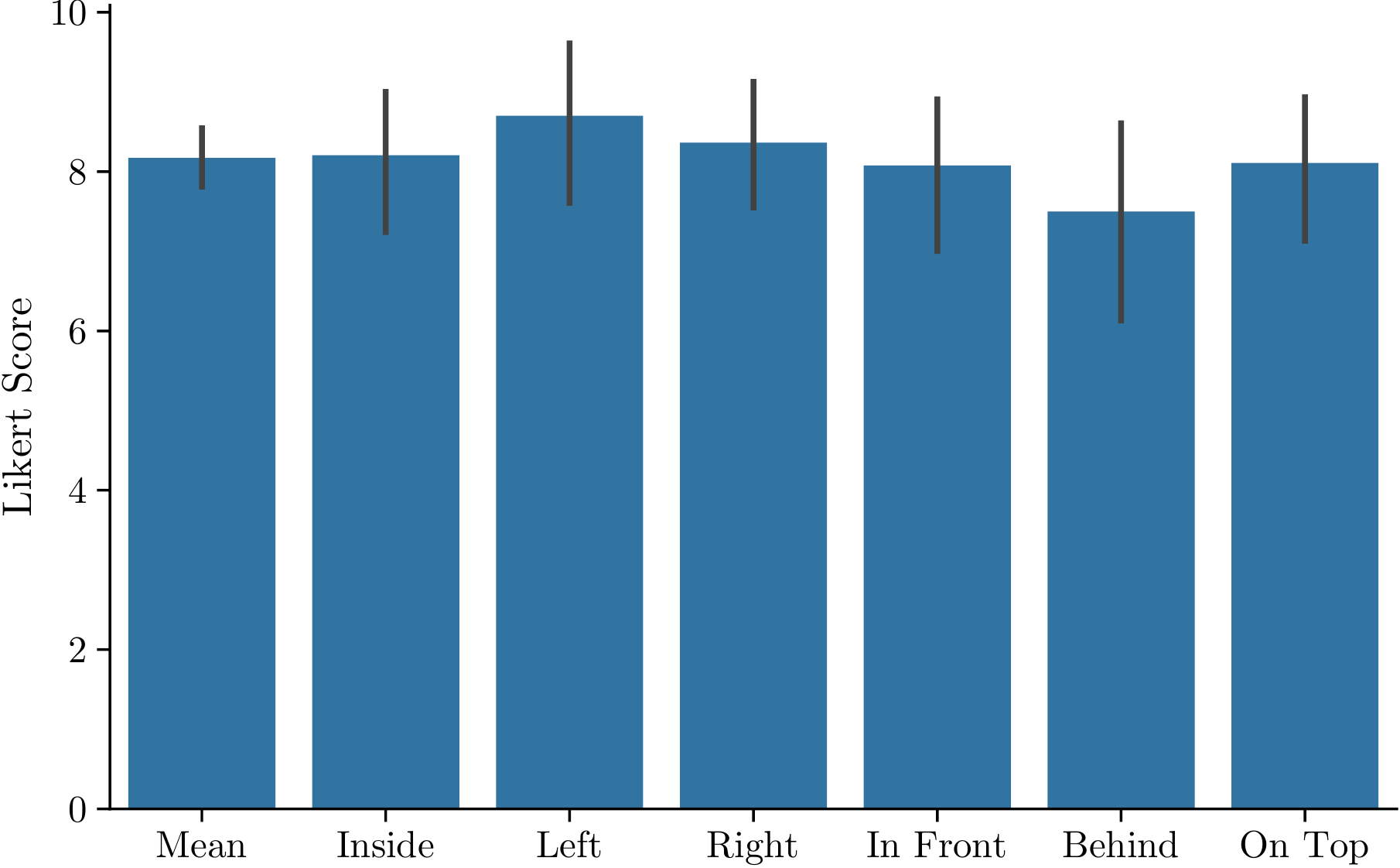}
   \caption{Performance of a PR2 robot following natural language instructions of 11 participants for object placement. Error bars indicate 95\% confidence intervals. }
\label{fig:user_ratings}
\end{figure}
\subsubsection{Procedure}
Our study involved 11 participants recruited from a university community. Each participant was asked to give 20 natural language instructions to the PR2 robot, which were parsed with an Amazon Echo Dot device. For each placement trial, the participants were asked to choose a reference item from a range of 30  household objects and to place it on the table at a random location. Next, the participants were instructed to choose a different item to put on the gripper of the robot. Afterwards, the participants were asked to provide a natural language instruction that contained one of the six spatial relations. We relied on keyword spotting to select the corresponding predicted distribution. The instruction was repeated in case of failures synthesizing the voice input. After sampling a $(u, v)$ location from the predicted distribution, we used the robot's Asus Xtion RGB-D camera to localize the pixel coordinate in 3D space. Our system then planned a top-down grasp pose to the calculated 3D point. The reachability of the proposed plan was checked using MoveIt!~\cite{chitta2012moveit}. The end-effector was moved above the desired location and then the gripper was opened to complete the placement. After each trial the participants rated the placement on a 10-point Likert scale as well as a binary success rate. The Likert scale helps us rate ambiguous placements such as a top left or diagonal placements for a instruction containing the relation ``left''.

\subsubsection{Results}
Figure \ref{fig:user_ratings} shows the performance of our approach on a PR2 robot for a total of 220 natural language instructions of 11 participants. We report a mean rating of 8.1 over all relations and trials. We observe a lower score for ``behind'' as many points sampled were outside the reach of the robot's gripper and therefore no valid motion plan was found. We observe a similar behaviour for the success rates reported on Table \ref{tab:sc}. Many participants chose reference items with a small area to place the object for the relations ``on top'' and ``inside'', requiring a precision placement. For such challenging scenarios, we observed some failure cases when the sampled location lied at the border of the reference item or the depth information from the RGB-D camera contained noise. Overall, our results demonstrate the ability of our approach to effectively learn object placements for relational instructions.
\begin{figure}
\setlength{\tabcolsep}{1.pt}
\begin{tabular}{ccc}
\includegraphics[width=0.33\linewidth, trim=0 0 40 90, clip]{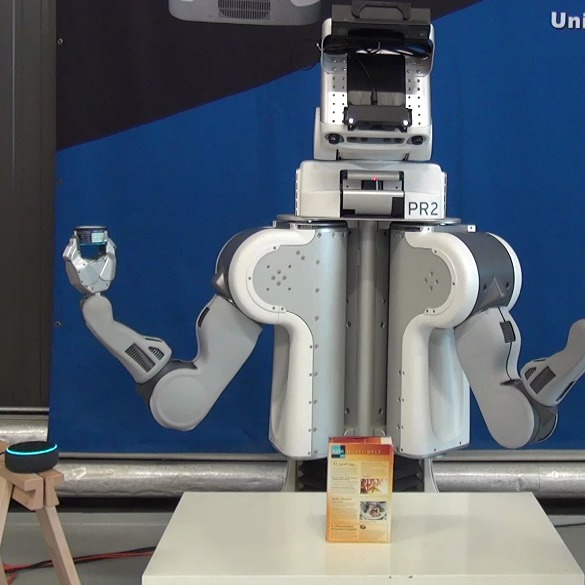}&
\includegraphics[width=0.33\linewidth, trim=0 0 40 90, clip]{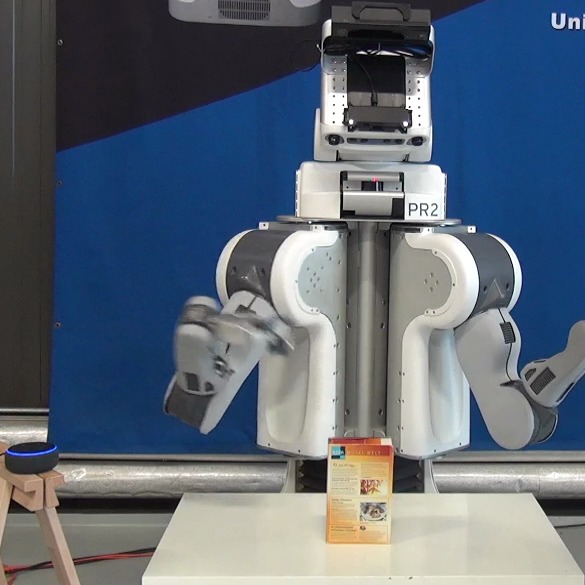}&
\includegraphics[width=0.33\linewidth, trim=0 0 40 90, clip]{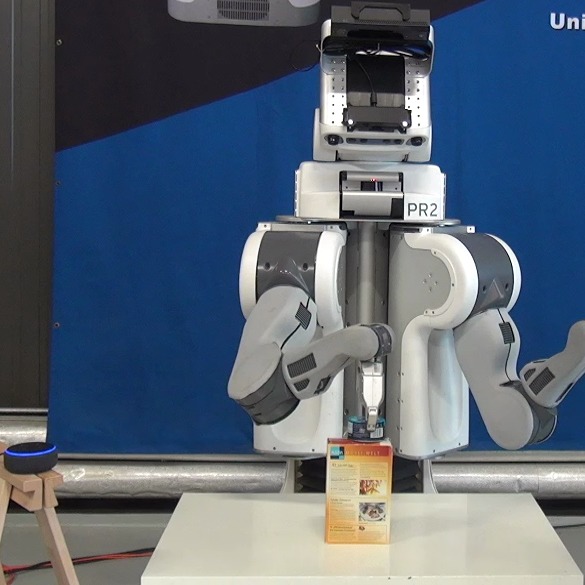}
\end{tabular}
\caption{Example object placing execution with the PR2 robot for the natural language instruction ``place the can on top of the box'', which is synthesized with the Amazon Echo Dot.} \label{fig:pr2grasp}
\end{figure}
\begin{table}[h]
 \setlength{\tabcolsep}{3.5pt}
  \begin{tabular}{l | c | c c c c c c}
  Metric    & Mean  & Inside & Left & Right & In Front & Behind & On Top\\
  \hline
  \hline
  Success Rate    & 0.84  & 0.84   & 0.87  & 0.95 & 0.84     & 0.80   & 0.79\\
  \end{tabular}
  \caption{Performance of our approach on a real robot platform following natural language instructions of 11 participants to place objects in a tabletop scenario.}
  \label{tab:sc}
\end{table} 
\vspace*{-2mm}
\section{Conclusion}
In this paper, we presented a novel approach to the problem of learning learning object placements for relational instructions from a single image. We  exemplified  how the distributions produced by our  method  enables a real-world robot to place objects by following relational natural language instructions.
Our method is based on leveraging three key ideas:~\emph{i}) 
modeling object-object spatial relations on natural images instead of 3D, which helps avoiding additional instrumentation for object tracking  and the need for large collection of corresponding 3D shapes and relational data ~\emph{ii}) reasoning about the best way to place objects to reproduce a spatial relation by estimating pixelwise, non-parametric distributions, without the use of priors~\emph{iii})
leveraging auxiliary learning to overcome the problem of unavailability of ground-truth pixelwise annotation of spatial relations and thus receive the training signal by classifying hallucinating scene representations. 

We feel that this is a promising first step towards enabling a shared understanding between humans and robots. In the future, we plan to extend our approach to incorporate understanding of referring expressions to develop a pick-and-place system that follows natural language instructions. 


\bibliographystyle{unsrt}
\bibliography{root}

\end{document}